\documentclass[10pt,journal,compsoc]{IEEEtran}
\usepackage{epsfig}
\usepackage{graphicx}
\newcommand\MYhyperrefoptions{bookmarks=true,bookmarksnumbered=true,
pdfpagemode={UseOutlines},plainpages=false,pdfpagelabels=true,
colorlinks=true,linkcolor={black},citecolor={black},urlcolor={black},
pdftitle={Bare Demo of IEEEtran.cls for Computer Society Journals},
pdfsubject={Typesetting},
pdfauthor={Michael D. Shell},
pdfkeywords={Computer Society, IEEEtran, journal, LaTeX, paper,
             template}}
\usepackage[\MYhyperrefoptions,pdftex]{hyperref}
\usepackage{amsmath}
\usepackage{amssymb}
\usepackage{booktabs}
\usepackage{makecell}
\usepackage{subfigure}
\usepackage{extarrows}
\usepackage{bm}
\usepackage{mathrsfs}
\newtheorem{Lemma}{Lemma}
\newtheorem{definition}{Definition}
\newtheorem{theorem}{Theorem}
\def\Sp{{\scriptsize{\textcircled{{\emph{\tiny{\textbf{Sp}}}}}}}}
\makeatletter
\DeclareRobustCommand\onedot{\futurelet\@let@token\@onedot}
\def\@onedot{\ifx\@let@token.\else.\null\fi\xspace}

\def\eg{\emph{e.g}\onedot} 
\def\ie{\emph{i.e}\onedot}

\def\wrt{w.r.t\onedot} 
\def\etal{\emph{et al}\onedot}
\makeatother

\usepackage[ruled]{algorithm2e}
\usepackage{algpseudocode}

\ifCLASSOPTIONcompsoc
  \usepackage[nocompress]{cite}
\else
  \usepackage{cite}
\fi
\ifCLASSINFOpdf
\else
\fi
\begin{document}

\title{Effective and Efficient Graph Learning for Multi-view Clustering}

\author{Quanxue~Gao,
        Wei~Xia,
        Xinbo~Gao,
        Xiangdong~Zhang,
        Qin~Li,
        and~Dacheng~Tao,~\IEEEmembership{Fellow,~IEEE}
\IEEEcompsocitemizethanks{
\IEEEcompsocthanksitem This work is supported by National Natural Science Foundation of China under Grants 61773302 and 61372069, Natural Science Basic Research Plan in Shaanxi Province (Grant 2020JZ-19), supported by the Innovation Fund of Xidian University.\protect

\IEEEcompsocthanksitem Corresponding author: Quanxue Gao, e-mail: qxgao@xidian.edu.cn.\protect

\IEEEcompsocthanksitem Q. Gao, W. Xia and X. Zhang are with the State Key laboratory of Integrated Services Networks, Xidian University, Xi’an 710071, China.\protect

\IEEEcompsocthanksitem Q. Li is with School of Software Engineering, Shenzhen Institute of Information Technology, Shenzhen 518172, China.\protect

\IEEEcompsocthanksitem X. Gao is with the School of Electronic Engineering, Xidian University, Xi’an 710071, China and with the Chongqing Key Laboratory of Image Cognition, Chongqing University of Posts and Telecommunications, Chongqing 400065, China.\protect

\IEEEcompsocthanksitem D. Tao is with the UBTECH Sydney Artificial Intelligence Centre and the School of Information Technologies, Faculty of Engineering and Information Technologies, University of Sydney, Darlington, NSW 2008, Australia. \protect}%
\thanks{Manuscript received December 13, 2020; revised ********; accepted ********.}}

\markboth{Journal of \LaTeX}%
{Shell \MakeLowercase{\textit{Gao et al.}}: Effective and Efficient Graph Learning for Multi-view Clustering}

\IEEEtitleabstractindextext{%
\begin{abstract}
Despite the impressive clustering performance and efficiency in characterizing both the relationship between data and cluster structure, existing graph-based multi-view clustering methods still have the following drawbacks. They suffer from the expensive time burden due to both the construction of graphs and eigen-decomposition of Laplacian matrix, and fail to explore the cluster structure of large-scale data. Moreover, they require a post-processing to get the final clustering, resulting in suboptimal performance. Furthermore, rank of the learned view-consensus graph cannot approximate the target rank. In this paper, drawing the inspiration from the bipartite graph, we propose an effective and efficient graph learning model for multi-view clustering. Specifically, our method exploits the view-similar between graphs of different views by the minimization of tensor Schatten $p$-norm, which well characterizes both the spatial structure and complementary information embedded in graphs of different views. We learn view-consensus graph with adaptively weighted strategy and connectivity constraint such that the connected components indicates clusters directly. Our proposed algorithm is time-economical and obtains the stable results and scales well with the data size. Extensive experimental results indicate that our method is superior to state-of-the-art methods.
\end{abstract}

\begin{IEEEkeywords}
Multi-view clustering, graph learning, tensor Schatten $p$-norm, connectivity constraint.
\end{IEEEkeywords}}

\maketitle
\IEEEdisplaynontitleabstractindextext
\IEEEpeerreviewmaketitle

\IEEEraisesectionheading{\section{Introduction}\label{sec:introduction}}
\IEEEPARstart{I}{n} real word applications, with the rapid development of sensor technology, each object can be usually sensed and described by different views. Drawing the inspiration from the principle that information embedded in different views are complementary and  convey the common underlying clusters, multi-view clustering has become an active topic in computer vision and pattern recognition~\cite{Our,XXGHXG,ZhangFHCXTX20,Jie2020,NieLL16,ZhanZGW18,HuNWL20}. Multi-view clustering aims to divide data into different groups such that the data in the same group have high similarity to each other, while data points in different groups have low similarity. Numerous clustering methods have been developed, among which
graph-based clustering is one of the most representative clustering techniques due to its efficiency in characterizing both the complex structure of data and relationship between data.


The purpose of graph-based multi-view clustering is to learn view-consensus graph by fusing graphs of different views. Two of the most representative methods are co-regularized multi-view spectral clustering (Co-reg)~\cite{KumarD11} and co-training multi-view spectral clustering (Co-train)~\cite{KumarRD11r}. Although Co-reg and Co-train have good performance, all of them treat all views equally, which does not make sense in real-word applications. To improve robustness of clustering algorithm, Nie \etal~\cite{NieLL16} adaptively assigned the weighted values for different views and developed auto-weighted graph learning method (AMGL). However, all of them need post-processing such as $k$-means to get clustering results, resulting in suboptimal performance. To solve this problem, Hu \etal integrated nonnegative embedding and spectral embedding into a unified framework, and proposed multi-view spectral clustering method (SMSC)~\cite{HuNWL20}. To reduce the computational complexity, Li \emph{et al}.~\cite{LiNHH15} developed a bipartite graph-based fast algorithm for multi-view spectral clustering (MVSC).

Although the motivations of the aforementioned methods are different, their performance heavily depends on the predefined graphs of different views. In real applications, it is very difficult to artificially design a suitable graph for each view. This reduces the flexibility of algorithms. To tackle this problem, Nie \emph{et al}.~\cite{NieCLL18} proposed multi-view learning with adaptive neighbours (MLAN) for clustering. It explicitly assumes that each data in different views has the same neighbors. This assumption is very strict, leading to suboptimal performance. To relax this assumption, Zhan \emph{et al}. ~\cite{ZhanZGW18} proposed graph learning for multi-view clustering (MVGL). Li \emph{et al}.~\cite{SFMC} proposed scalable and parameter-free multi-view clustering (SFMC), which is effective and efficient for large scale multi-view clustering.

However, the aforementioned methods cannot well exploit the complementary information and spatial structure of graph. Most algorithms have demonstrated that low-rank representation can well characterize the relationship between data and exploit the spatial structure, thus graph-based multi-view clustering with low-rank constraint has attracted more and more attention. One of the most representative methods is robust multi-view spectral clustering (RMSC). It learns the view-consensus graph, which is shared by different views, via the nuclear norm minimization. But it cannot well exploit the complementary information and high-order information embedded in graphs of different views. To handle this problem, inspired by the recently proposed tensor nuclear norm, Wu \emph{et al}.~\cite{WuLZ19} presented essential tensor learning for multi-view spectral clustering (ETLMSC). Although impressive clustering performance, ETLMSC still has the following limitations. (1) It is high time-consuming due to both the graph construction and eigen-decomposition of Laplacian matrix. The computational complexity of ETLMSC is $\bm{\mathcal{O}}(Vn^2d)$ for graph construction and $\bm{\mathcal{O}}(n^3)$ for eigen-decomposition of Laplacian matrix, respectively, where $V$, $n$ and $d$ denote the number of views, data and feature dimensionality, respectively. It is easy to see that ETLMSC fails to deal with large scale datasets, which is one of the most important problems in real applications~\cite{XuHNL17,LiNHH15}.
(2) The learned graph does not well characterize the cluster structure. Thus, it requires post-processing to get clustering results, resulting in suboptimal clustering performances. (3) Rank of the learned graph cannot well approximate the target rank. So it cannot well exploit the complementary information embedded in different views.

\begin{figure}[!t]
\begin{center}
\includegraphics[width=1.0\linewidth]{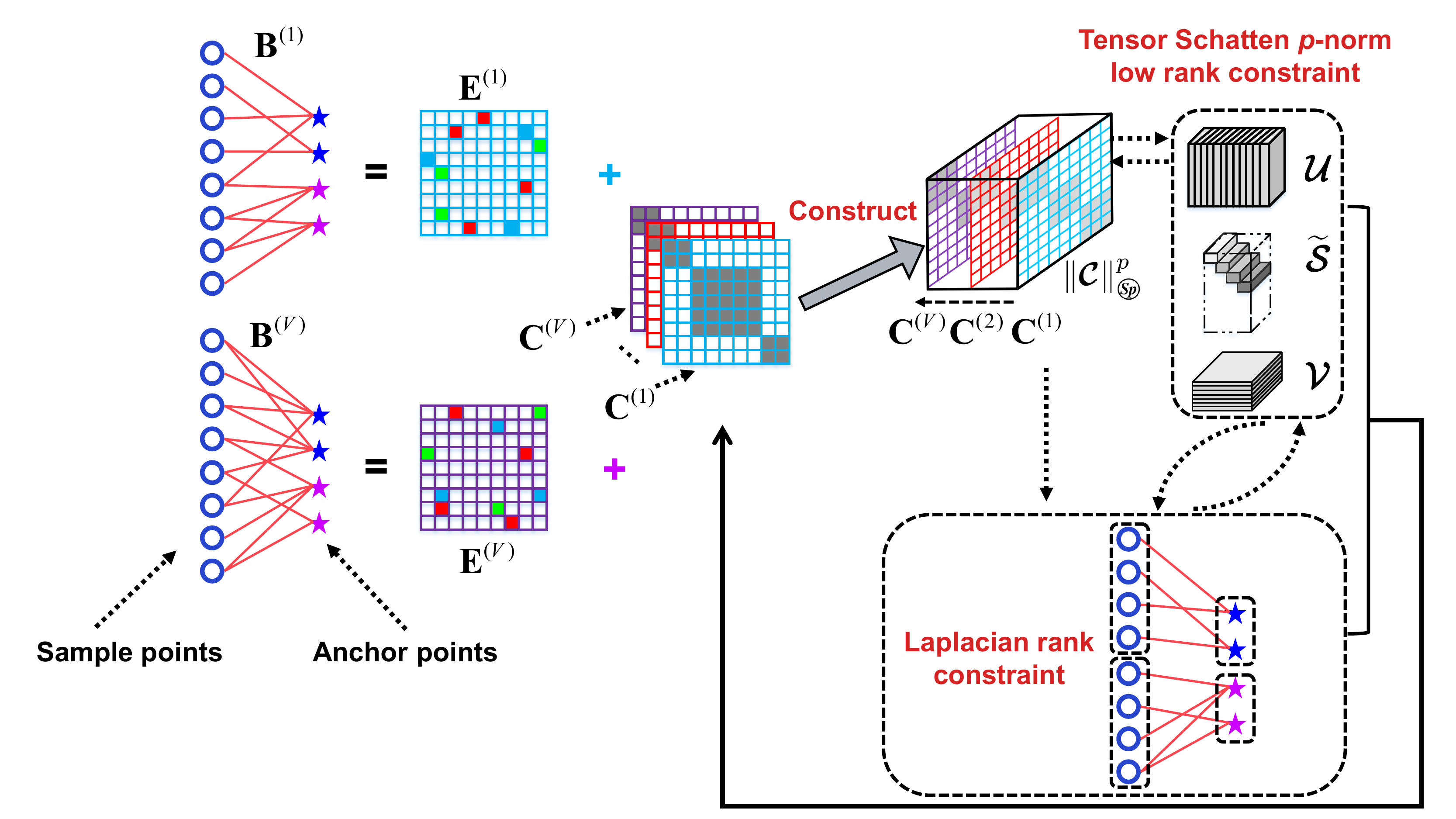}
\end{center}
   \caption{The flowchart of our method. $\mathbf{X}^{(v)}$ and $\mathbf{B}^{(v)}$ are data matrix and graphs of $v$-th view, respectively. $\mathbf{E}^{(v)}$ is error.}
\label{flowchart}
\vspace{-5mm}
\end{figure}

To handle the aforementioned problems, drawing inspiration from bipartite graph~\cite{LiuHC10} and Schatten $p$-norm~\cite{NieHD12}, we propose a scalable graph learning model, which can be applied to large scale multi-view clustering (See Fig.~\ref{flowchart}). Specifically, to avoid constructing the $n \times n$ graph, we construct a bipartite graph whose size is $n\times m$ ($m\ll n$), where $m$ denotes the number of anchors, and then leverage our proposed tensor Schatten $p$-norm to exploit the view-similar embedded in graphs in different graphs. To well exploit the cluster structure, we learn view-consensus graph with adaptively weighted strategy and connectivity constraint such that the connected components indicates clusters directly. Finally, we proposed an efficient and fast algorithm to solve eigen-decomposition of Laplacian matrix. Thus, for large scale multi-view databases, our method remarkably reduce the complexity from $\bm{\mathcal{O}}(n^3 + Vn^2d)$ to $\bm{\mathcal{O}}(m^2n+Vnmd)$. The contributions are summarized as follows:
\begin{itemize}
  \item We employ the minimization of tensor Schatten $p$-norm, which helps get the target rank, to exploit the view-similar between graphs of different views. Thus, the rank of the learned view-consensus graph well approximates the target rank.
  \item The learned view-similar graph has $K$-connected components and well adaptively takes into account importance of different views. Our method directly gets the clustering results according the connected components without any post-processing.
  \item Our proposed algorithm reduces the main computational complexity from $\bm{\mathcal{O}}(n^3 + Vn^2d)$ to $\bm{\mathcal{O}}(m^2n+Vnmd)$, compared with ETLMSC. Thus, our proposed algorithm is time-economical and well suitable for large-scale multi-view data clustering.
\end{itemize}

\textbf{\emph{Notations.}} In this paper, we use bold calligraphy letters for third-order tensors, \eg, ${\bm{\mathcal {D}}} \in{\mathbb{R}} {^{{n_1} \times {n_2} \times {n_3}}}$, bold upper case letters for matrices, \eg, ${\bf{D}}$, bold lower case letters for vectors, \eg, ${\bf{d}}$, and lower case letters such as ${d_{ijk}}$ for the entries of ${\bm{\mathcal {D}}}$. The $i$-th frontal slice of ${\bm{\mathcal {D}}}$ is ${\bm{\mathcal {D}}}^{(i)}$. $\overline {{\bm{\mathcal {D}}}}$ is the discrete Fast Fourier Transform (FFT) of ${\bm{\mathcal {D}}}$ along the third dimension, \ie, $\overline {{\bm{\mathcal {D}}}} = fft({{\bm{\mathcal D}}},[ ],3)$. Thus, $\bm{{\mathcal D}} = ifft({\overline {\bm{\mathcal D}}},[ ],3)$. The trace of matrix $\mathbf{D}$ is denoted by $tr(\mathbf{D})$. The $\ell_{1}$-norm of $\mathbf{D}$ is written as $\left\| \mathbf{D} \right\| _1$. $\mathbf{I}$ is an identity matrix.

\section{Methodology}
\subsection{Problem Formulation and Objective}
Let $\{{\bf{X}}^{(v)}\}_{v=1}^V$ and $\{{\bf{G}}^{(v)}\}_{v=1}^V$ denote the data matrix and graph of the $v$-th view respectively, where ${{\bf{X}}^{(v)}} \in{\mathbb{R}} {^{n\times {d_v}}}$, ${{\bf{G}}^{(v)}} \in{\mathbb{R}} {^{n\times {n}}}$, ${d_v}$ and $n$ denote the number of dimensions and data points in the $v$-th view respectively, $V$ denotes the view number of data. The objective of ETLMSC is 
\begin{equation}\label{ETLMSC}
\begin{aligned}
\underset{\bm{\mathcal{D}},\bm{\mathcal{E}}}{\min}\| \bm{\mathcal{D}} \| _{\circledast}+\alpha \| \bm{\mathcal{E}} \| _{2,1}\,\, \quad s.t. \quad \bm{\mathcal{G}}=\bm{\mathcal{D}}+\bm{\mathcal{E}}
\end{aligned}
 \end{equation}
where $\bm{\mathcal{G}} (:,v,:) = {\bf{G}}^{(v)}$, $\bm{\mathcal{D}}$ denotes clear graph, $\bm{\mathcal{E}}$ is error matrix. $\| {\bm{\mathcal{D}}} \|_{\circledast}={\sum\nolimits_{v = 1}^{{V}} {\| {{{\overline {\bf{D}}}^{(v)}}} \|}_*}$ is the tensor nuclear norm (TNN) of $\bm{\mathcal{D}}$, ${\| {{{\overline {\bf{D}} }^{(v)}}}\|}_* $ denotes nuclear norm of ${{\overline {\bf{D}} }^{(v)}} \in {\mathbb{R}} {^{{n} \times {V}}} $, which is the sum of all singular values of ${{\overline {\bf{D}} }^{(v)}}$, $\alpha$ is a trade-off parameter.




After obtained the graphs ${\bf{D}}^{(v)}$ ($v=1,2,\cdots V$), it needs to leverage standard spectral clustering to get the final clustering. Thus, in the model (\ref{ETLMSC}), the main time of algorithm is spent on the graph construction ${\bf{G}}^{(v)} \in{\mathbb{R}} {^{n\times n}}$ ($v=1,2,\cdots, V$) and eigen-decomposition of Laplacian matrix ${\bf{L}} \in{\mathbb{R}} {^{n\times n}}$. Their computational complexity are $\bm{\mathcal{O}}(Vn^2d)$ and $\bm{\mathcal{O}}(n^3)$, respectively. It is easy to see that this model suffers from the expensive time burden and fails to deal with large scale database, which is usually encountered in real-world applications. Moreover, the learned graph does not have $K$-connected components, so it needs post-processing to get clustering results, leading to the sub-optimal graph and clustering performance. Finally, the rank of the learned graph does not approximate the target rank. For clustering, an ideal similarity graph should have low-rank structure and $K$-connected components, where $K$ is the cluster number, \ie, the target rank. Since tensor nuclear norm minimization does not guarantee that the rank of clean tensor approximates the target rank. The learned graph, which is obtained by the model (\ref{ETLMSC}), does not characterize the cluster structure.

To handle the aforementioned problems, we propose an effective and efficient model for multi-view clustering. Specifically, to remarkably reduce the computational complexity, drawing the inspiration from bipartite graph~\cite{LiuHC10}, we construct an effective bipartite graph ${\bf{B}}^{(v)} \in{\mathbb{R}} {^{n\times m}}$, which exploits the relationship between $n$ data points and $m$ ($m\ll n$) anchors, instead of $n\times n$ global graph ${\bf{G}}^{(v)}$. This remarkably reduces the main computational complexity from $\bm{\mathcal{O}}(Vn^2d)$ to $\bm{\mathcal{O}}(Vnmd)$. Meanwhile, inspired by Lemma~\ref{lem2}, we leverage the Laplacian rank constraint to ensure that the learned graphs ${\bf{C}}^{(v)}$ have $K$-connected components. This characterizes the cluster structure of data and helps get the cluster results without any post-processing. 
 Thus, we have 
\begin{equation}\label{f-1}
\begin{aligned}
\mathop {\min }\limits_{\scriptstyle{\bf{C}}^{(v)}{\bf{1}} = {\bf{1}},{{\bf{C}}^{(v)}} \ge 0,{\bf{E}}^{(v)}\hfill}\| \bm{\mathcal{C}} \| _{\circledast}+ \alpha \sum\limits_{v = 1}^V{\| \mathbf{E}^{( v )} \| _1}\quad\quad\\
\text {s.t.}\quad\mathbf{B}^{( v )}=\mathbf{C}^{( v )}+\mathbf{E}^{( v )}, rank(\widetilde{\mathbf{L}}_{\mathbf{F}^{(v)}})=n+m-K\\
\end{aligned}
\end{equation}
where  ${\bm{{\mathcal C}}}(:,v,:) = {{\bf{C}}^{(v)}}$,
$\widetilde{\mathbf{L}}_{\mathbf{F}^{(v)}}=\mathbf{I}-\mathbf{D}_{\mathbf{F}^{(v)}}^{-\frac{1}{2}}{\mathbf{F}^{(v)}}\mathbf{D}_{\mathbf{F}^{(v)}}^{-\frac{1}{2}}$ is the normalized Laplacian matrix of ${\mathbf{F}^{(v)}} \in \mathbb{R}^{(n+m)\times (n+m)}$ with ${\mathbf{F}^{(v)}}=\left[ \begin{matrix}&\mathbf{C}^{(v)}\\ \mathbf{C}^{(v)^{\mathbf{T}}}&\\\end{matrix} \right]$. 
 $\mathbf{D}_{\mathbf{F}^{(v)}}$ is a diagonal matrix whose diagonal elements are $\mathbf{D}_{\mathbf{F}^{(v)}}\left( i,i \right) =\sum\nolimits_{j=1}^{n+m}{f_{{ij}}^{(v)}}$.
Since ${\mathbf{F}^{(v)}}$ is intrinsically comprised of double $\mathbf{C}^{(v)}$, the $K$-connected ${\mathbf{F}^{(v)}}$ certainly guarantees the $K$-connected $\mathbf{C}^{(v)}$. 

\begin{Lemma}~\cite{LL1}\label{lem2} The multiplicity $K$ of the eigenvalue zeros of $\widetilde{\mathbf{L}}_{\mathbf{F}^{(v)}}$ is equals to the number of connected components in the graph associated with ${\mathbf{C}^{(v)}}$.
\end{Lemma}

In the model (\ref{f-1}), we leverage the eigen-value minimization to approximate the rank constraint which is a non-convex problem and hard to tackle. Inspired by the Ky Fan's Theorem~\cite{KFF}, the constraint $rank(\widetilde{\mathbf{L}}_{\mathbf{F}^{(v)}})=n+m-K$ can be approximated by solving the model (\ref{f-3}).
\begin{equation}\label{f-3}
\begin{aligned}
\mathop {\min }\limits_{\scriptstyle{{\mathbf{{P}}^{(v)}}^\mathbf{T}}{\mathbf{{P}}^{(v)}} = {\bf{I}}\hfill}& \sum\limits_{v = 1}^Vtr( \mathbf{P}^{{(v)^\mathbf{T}}}\widetilde{\mathbf{L}}_{\mathbf{F}^{(v)}}\mathbf{P}^{(v)} )\\
\end{aligned}
\end{equation}
where $\mathbf{P}^{(v)}=[\mathbf{p}_1^{(v)};\cdots;\mathbf{p}_{n+m}^{(v)}]\in \mathbb{R}^{(n+m)\times K}$ is the indicator matrix of the $v$-th view.

For multi-view data, each view contains some content of the object that other views do not. Thus, graphs, which are constructed from different views, usually has significant different role for clustering. However, Eq. (\ref{f-3}) ignores this. To improve robustness of our model, we rewrite Eq. (\ref{f-3}) as
\begin{equation}\label{f-add-1}
\begin{aligned}
\mathop {\min }\limits_{\scriptstyle{{\bf{P}}^\mathbf{T}}{\bf{P}} = {\bf{I}}\hfill \atop \scriptstyle{{\xi ^{(v)}}}\hfill} \sum\limits_{v = 1}^V tr({\frac{1}{\xi ^{(v)}}} \mathbf{P}^{{\mathbf{T}}}\widetilde{\mathbf{L}}_{\mathbf{F}^{(v)}}\mathbf{P}),\text {s.t.}\ {\sum\limits_{v = 1}^V {{\xi ^{(v)}} = 1} ,{\xi ^{(v)}} \ge 0}
\end{aligned}
\end{equation}

Then, the model (\ref{f-1}) can be rewritten as
\begin{equation}\label{f-p}
\begin{aligned}
&\mathop {\min }\limits_{\scriptstyle{{\bf{C}}^{(v)}}, {\bf{E}}^{(v)}, {\bf{P}}, {\xi ^{(v)}}\hfill}\| \bm{\mathcal{C}} \| _{\circledast}+ \alpha \sum\limits_{v = 1}^V{\| \mathbf{E}^{( v )} \| _1}+ \beta tr(\mathbf{P}^{{\mathbf{T}}}\widetilde{\mathbf{L}}\mathbf{P})\\
&\quad\text {s.t.}\quad \mathbf{B}^{( v )}=\mathbf{C}^{( v )}+\mathbf{E}^{( v )}, {\sum\limits_{v = 1}^V {{\xi ^{(v)}} = 1} ,{\xi ^{(v)}} \ge 0}\\
&\quad \quad \quad \quad {\kern 1pt}{\kern 1pt}{\kern 1pt} {\bf{C}}^{(v)}{\bf{1}} = {\bf{1}},{{\bf{C}}^{(v)}} \ge 0, {{\bf{P}}^\mathbf{T}}{\bf{P}} = {\bf{I}}
\end{aligned}
\end{equation}
where $\widetilde{\mathbf{L}}=\sum_{v=1}^V{\frac{1}{\xi ^{(v)}}}\widetilde{\mathbf{L}}_{\mathbf{F}^{(v)}}$.

To ensure that the rank of ${\bm{{\mathcal C}}}$ approximates the target rank, inspired by Schatten $p$-norm, we present tensor Schatten $p$-norm (See Definition \ref{definitionTSP}) and use it instead of the first term in the model (\ref{f-p}).

\begin{definition}\label{definitionTSP}
Given ${\bm{\mathcal {X}}} \in{\mathbb{R}} {^{{n_1} \times {n_2} \times {n_3}}}$, $h ={\min ({n_1},{n_2})}$, tensor Schatten $p$-norm of tensor ${\bm{\mathcal {X}}}$ is defined as
\begin{equation}\label{eq3}
\begin{aligned}
{\| \bm{{\mathcal X}} \|_\Sp}=(\sum\limits_{i = 1}^{n3} {\| {{{{{\overline {\bm{\mathcal {X}}}}}}^{(i)}}} \|_\Sp^p{)^{\frac{1}{p}}}} = ({\sum\limits_{i = 1}^{n3} {\sum\limits_{j = 1}^{h} {{\sigma _j}{{({{{{\overline {\bm{\mathcal {X}}}}}}^{(i)}})^p}}} )^{\frac{1}{p}}}}
\end{aligned}
 \end{equation}
where, $0<p\leq 1$, ${\sigma _j}({{{\overline {\bm{\mathcal {X}}}}}^{(i)}})$ denotes the $j$-th singular value of ${{{\overline {\bm{\mathcal {X}}}}}^{(i)}}$.
\end{definition}

Finally, our objective function is
\begin{equation}\label{f}
\begin{aligned}
&\mathop {\min }\limits_{\scriptstyle{{\bf{C}}^{(v)}}, {\bf{E}}^{(v)}, {\bf{P}}, {\xi ^{(v)}}\hfill}\| \bm{\mathcal{C}} \| _{\Sp}^p+ \alpha \sum\limits_{v = 1}^V{\| \mathbf{E}^{( v )} \| _1}+ \beta tr(\mathbf{P}^{{\mathbf{T}}}\widetilde{\mathbf{L}}\mathbf{P})\\
&\quad\text {s.t.}\quad \mathbf{B}^{( v )}=\mathbf{C}^{( v )}+\mathbf{E}^{( v )}, {\sum\limits_{v = 1}^V {{\xi ^{(v)}} = 1} ,{\xi ^{(v)}} \ge 0}\\
&\quad \quad \quad \quad {\kern 1pt}{\kern 1pt}{\kern 1pt} {\bf{C}}^{(v)}{\bf{1}} = {\bf{1}},{{\bf{C}}^{(v)}} \ge 0, {{\bf{P}}^\mathbf{T}}{\bf{P}} = {\bf{I}}
\end{aligned}
\end{equation}

\subsection{Optimization}

Inspired by augmented Lagrange multiplier (ALM)~\cite{LinLS11}, we introduce an auxiliary variable $\bm{{\mathcal J}}$ and rewrite the model (\ref{f}) as the following unconstrained problem:
\begin{equation}\label{o1}
\begin{aligned}
\bm{{\mathcal L}}&({{\bf{C}}^{(1)}}, \cdots,{{\bf{C}}^{(m)}}, \bm{{\mathcal J}},{{\mathbf{E}}^{(1)}}, \cdots,{{\bf{E}}^{(m)}}, {\bf{P}})\\
&= { \| \bm{{\mathcal J}} \|_{\Sp}^p} + \alpha \sum\limits_{v = 1}^V{\| \mathbf{E}^{( v )} \| _1} + \beta tr( \mathbf{P}^\mathbf{T}\widetilde{\mathbf{L}}\mathbf{P} )\\
& \ \ \ \ +\sum\limits_{v = 1}^V ( {\langle {{{\bf{Y}}_1^{(v)}},\mathbf{B}^{( \mathrm{v} )}-\mathbf{E}^{( \mathrm{v} )}-\mathbf{C}^{( \mathrm{v} )}} \rangle }
\\
&\qquad \ \ \ \ \ \ +\frac{\mu }{2}\| \mathbf{B}^{( v )}-\mathbf{E}^{( v )}-\mathbf{C}^{( v )}\|_F^2 )\\
& \ \ \ \ +\langle {\bm{{\mathcal Y}}_2,\bm{{\mathcal C}} - \bm{{\mathcal J}}} \rangle  + \frac{\rho }{2}\| {\bm{{\mathcal C}} - \bm{{\mathcal J}}} \|_F^2
\end{aligned}
\end{equation}
where ${{\bf{Y}}_1}^{(v)}$ and $\bm{{\mathcal Y}}_2$ represent Lagrange multipliers, $\mu $ and $\rho $ are the penalty parameters. Consequently, the optimization process could be separated into four steps:

$\bullet$ \textbf{Solving ${\bf{P}}$ with fixed ${\bf{C}}^{(v)}$, ${\bf{E}}^{(v)}$, ${\xi}^{(v)}$ and $\bm{{\mathcal J}}$.} In this case, the optimization \wrt ${\bf{P}}$ in Eq. (\ref{o1}) becomes
\begin{equation}\label{P-2}
\begin{aligned}
\mathbf{P}^{\ast}&=\mathop {\arg \min }\limits_{{{\bf{P}}^\mathbf{T}}{\bf{P}} = {\bf{I}}} tr(\mathbf{P}^\mathbf{T}\widetilde{\mathbf{L}}\mathbf{P})\\
\end{aligned}
\end{equation}

To directly optimize the model (\ref{P-2}), the computational complexity is $\bm{\mathcal{O}}((n+m)^2K)$. Instead of it, we herein provide an efficient fast algorithm. Substituting $\widetilde{\mathbf{L}}=\sum\limits_{v=1}^V{\frac{1}{\xi ^{(v)}}}\left(\mathbf{I}-\mathbf{D}_{{\mathbf{F}}^{(v)}}^{-\frac{1}{2}}\mathbf{F}^{(v)}\mathbf{D}_{{\mathbf{F}}^{(v)}}^{-\frac{1}{2}}\right)$ into Eq. (\ref{P-2}), we have
\begin{equation}\label{P-3}
\begin{aligned}
\mathbf{P}^{\ast}=\mathop {\arg \min }\limits_{{{\bf{P}}^\mathbf{T}}{\bf{P}} = {\bf{I}}} tr(\mathbf{P}^\mathbf{T}\sum\limits_{v=1}^V{\frac{1}{\xi ^{(v)}}}\mathbf{P}-\mathbf{P}^\mathbf{T}\mathbf{D}_{{\mathbf{F}}^{(v)}}^{-\frac{1}{2}}\mathbf{F}^{(v)}\mathbf{D}_{{\mathbf{F}}^{(v)}}^{-\frac{1}{2}}\mathbf{P})\\
\end{aligned}
\end{equation}

Let $\mathbf{P}=\left[ \begin{array}{c}	\mathbf{P}_U\\	\mathbf{P}_M\\ \end{array} \right] $ and $\mathbf{D}_\mathbf{{F}}^{(v)}=\left[ \begin{matrix} 	\mathbf{D}_U^{(v)}&		\\	&		\mathbf{D}_M^{(v)}\\ \end{matrix} \right] $, where $\mathbf{P}_U \in \mathbb{R}^{n\times K}$ is the first $n$ rows of $\mathbf{P}$ and $\mathbf{P}_M \in \mathbb{R}^{m\times K} $ is the remaining $m$ rows of $\mathbf{P}$, $\mathbf{D}_U^{(v)} \in \mathbb{R} ^{n \times n}$ and $\mathbf{D}_M^{(v)} \in \mathbb{R} ^{m \times m}$ are diagonal matrices whose diagonal elements are $\mathbf{D}_U^{(v)}(i,i)=\sum\nolimits_{j=1}^m{{\bf{C}}^{(v)}(i,j)}$ and $\mathbf{D}_M^{(v)}(j,j)=\sum\nolimits_{i=1}^n{{\bf{C}}^{(v)}(i,j)}$. Substituting them into Eq. (\ref{P-3}), and by simple algebra, Eq. (\ref{P-3}) becomes
\begin{equation}\label{P-1}
\begin{aligned}
\mathbf{P}^{\ast}=\mathop {\arg \max }\limits_{{\mathbf{P}_U^{\mathbf{T}}}{\mathbf{P}_U} + {\mathbf{P}_M^{\mathbf{T}}}{\mathbf{P}_M} = {\bf{I}}} 2tr(\mathbf{P}_U^{\mathbf{T}}\sum_{v=1}^V{\frac{\mathbf{C}^{(v)}{\mathbf{D}_{M}^{(v)}}^{-\frac{1}{2}}}{\xi ^{(v)}}}\mathbf{P}_M)
\end{aligned}
\end{equation}

Let $\mathbf{W}=\sum_{v=1}^V{\frac{\mathbf{C}^{(v)}{\mathbf{D}_{M}^{(v)}}^{-\frac{1}{2}}}{\xi ^{(v)}}}$. The optimal $\mathbf{P}^{\ast}$ in Eq. (\ref{P-1}) can be solved by Lemma~\ref{lem1}~\cite{NieWDH17}.
\begin{Lemma}\label{lem1} Suppose $\mathbf{W}\in \mathbb{R}^{n\times m}$, $\mathbf{P}_U\in \mathbb{R}^{n\times K}$, $\mathbf{P}_M\in \mathbb{R}^{m\times K}$. The optimal solutions to the following model:
\begin{equation}\label{P-1-1}
\max_{{\mathbf{P}_U^{\mathbf{T}}}{\mathbf{P}_U} + {\mathbf{P}_M^{\mathbf{T}}}{\mathbf{P}_M} = {\bf{I}}} tr(\mathbf{P}_U^{\mathbf{T}}\mathbf{W}\mathbf{P}_M )
\end{equation}
are $\mathbf{P}_U=\frac{\sqrt{2}}{2}\mathbf{U}_1$, $\mathbf{P}_M=\frac{\sqrt{2}}{2}\mathbf{V}_1$, where $\mathbf{U}_1$ and $\mathbf{V}_1$ are the leading $K$ left and right singular vectors of $\mathbf{W}$, respectively.
\end{Lemma}

Consequently, the optimal $\mathbf{P}^{\ast}=\frac{\sqrt{2}}{2}\left[ \begin{array}{c}	\mathbf{U}_1\\	\mathbf{V}_1\\ \end{array} \right]$, where $\mathbf{U}_1$ and $\mathbf{V}_1$ could be obtained by performing SVD decomposition on $\mathbf{W}$, which takes the computational complexity $\bm{\mathcal{O}}(Vnm + m^2n)$. Due to the number of anchors $m\ll n$ toward large-scale clustering, it is much more efficient to directly solve Eq. (\ref{P-2}) by tackling Eq. (\ref{P-1}) instead.

$\bullet$ \textbf{Solving ${\bf{C}}^{(v)}$ with fixed ${\bf{E}}^{(v)}$, $\bm{{\mathcal J}}$, ${\xi}^{(v)}$ and ${\bf{P}}$.} Now, the optimization \wrt ${\bf{C}}^{(v)}$ in Eq. (\ref{o1}) is equivalent to
\begin{equation}\label{t6}
\begin{aligned}
\mathop {\min }\limits_{\scriptstyle{\bf{C}}^{(v)}\hfill} &\langle {{{{\bf{Y}}_2^{(v)}}},{{{\bf{C}}^{(v)}}} - {{{\bf{J}}^{(v)}}}} \rangle + \frac{\rho }{2}\| {{{\bf{C}}^{(v)}}} - {{{\bf{J}}^{(v)}}} \|_F^2\\
&+{\langle {{{\bf{Y}}_1^{(v)}},\mathbf{B}^{( v )}-\mathbf{E}^{( v )}-\mathbf{C}^{( v )}} \rangle }\\
&+\frac{\mu }{2}\| \mathbf{B}^{( v )}-\mathbf{E}^{( v )}-\mathbf{C}^{( v )}
\|_F^2 + \beta tr( \mathbf{P}^\mathbf{T}\widetilde{\mathbf{L}}\mathbf{P} )\\
=&\mathop {\min }\limits_{\scriptstyle{\bf{C}}^{(v)}\hfill} { {\frac{\mu }{2}\| {\mathbf{Q}^{( v )}-\mathbf{C}^{( v )} } \|_F^2}}\\
&\quad \ \ \ +{\frac{\rho }{2}\| {{{\bf{C}}^{(v)}} - {{\bf{G}}^{(v)}}} \|_F^2}+\beta tr( \mathbf{P}^\mathbf{T}\widetilde{\mathbf{L}}\mathbf{P} )\\
&\quad\text {s.t.}\quad{\kern 1pt}{\kern 1pt}{\kern 1pt} {\bf{C}}^{(v)}{\bf{1}} = {\bf{1}},{{\bf{C}}^{(v)}} \ge 0
\end{aligned}
\end{equation}
where $\mathbf{G}^{( v )}=\mathbf{J}^{( v )}-\frac{1}{\rho}\mathbf{Y}_{2}^{( v )}$, $\mathbf{Q}^{( v )}=\mathbf{B}^{( v )}-\mathbf{E}^{( v )}+\frac{1}{\mu} \mathbf{Y}_{1}^{( v )}$.

According to Eq. (\ref{P-1}), the last term in Eq. (\ref{t6}) can be rewritten as
\begin{equation}\label{t8}
\begin{array}{l}
tr(\mathbf{P}^\mathbf{T}\widetilde{\mathbf{L}}\mathbf{P})=tr(\mathbf{P}^\mathbf{T}\sum\limits_{v=1}^V{\frac{1}{\xi ^{(v)}}}\mathbf{P})-2\sum\limits_{v=1}^V{tr(\mathbf{C}^{(v)^{\mathbf {T}}}{\mathbf{H}^{(v)^\mathbf{T}}} )}
\end{array}
\end{equation}
where ${\mathbf{H}^{(v)}}^\mathbf{T}={\frac{{\mathbf{D}_{M}^{(v)}}^{-\frac{1}{2}}}{\xi ^{(v)}}}\mathbf{P}_M\mathbf{P}_U^{\mathbf{T}}$. Thus, minimizing the Eq. (\ref{t6}) is equivalent to
\begin{equation}\label{t9}
\begin{aligned}
\mathop {\min }\limits_{\scriptstyle{\bf{C}}^{(v)}\hfill} &\frac{\rho + \mu}{2}tr(( \mathbf{C}^{( v )} ) ^\mathbf{T}\mathbf{C}^{( v )})- \rho tr(( \mathbf{C}^{( v )} ) ^\mathbf{T}\mathbf{G}^{( v )})\\
&-\mu tr(\mathbf{C}^{(v)^{\mathbf {T}}}\mathbf{Q}^{( v )})- 2{\beta}{tr( \mathbf{C}^{(v)^{\mathbf {T}}}{\mathbf{H}^{{(v)}^\mathbf{T}}} )}\\
=&\mathop {\min }\limits_{\scriptstyle{\bf{C}}^{(v)}\hfill}\frac{\rho + \mu}{2}\| \mathbf{C}^{( v )}-\frac{\bm{\varLambda}}{\rho +\mu} \| _{F}^{2}\\
&\quad\text {s.t.}\quad  {\kern 1pt}{\kern 1pt}{\kern 1pt} {\bf{C}}^{(v)}{\bf{1}} = {\bf{1}},{{\bf{C}}^{(v)}} \ge 0
\end{aligned}
\end{equation}
where $\bm{\varLambda}=\rho \mathbf{G}^{( v )}+\mu \mathbf{Q}^{( v )}+ 2 \beta {\mathbf{H}^{(v)^\mathbf{T}}}$. To this end, the closed-form solution ${{\bf{C}}^{(v)^{\ast}}}$ is ~\cite{NieWJH16} $\mathbf{c}_{i}^{( v )^{\ast}}=( \frac{\bm{\varLambda} _i}{\rho +\mu}+\gamma \mathbf{1} ) _+$, where $\gamma$ is the Lagrangian multiplier.

$\bullet$ \textbf{Solving ${\bf{E}}^{(v)}$ with fixed ${\bf{C}}^{(v)}$, $\bm{{\mathcal J}}$, ${\bf{P}}$ and ${\xi}^{(v)}$.} In this case, the optimization \wrt ${\bf{E}}^{(v)}$ in Eq. (\ref{o1}) becomes
\begin{equation}\label{E1}
\begin{aligned}
\mathbf{E}^{( v ) ^{\ast}}=\mathop {\mathrm{arg}\min}_{\mathbf{E}^{( v )}}\frac{\alpha}{\mu}\| \mathbf{E}^{( v )} \| _1+\frac{1}{2}\| \mathbf{E}^{( v )}-\Theta^{( v )} \| _{F}^{2}
\end{aligned}
\end{equation}
where $\Theta^{( v )}$ is $\Theta^{( v )}=\mathbf{B}^{( v )}-\mathbf{C}^{( v )}+ \frac{1}{\mu}\mathbf{Y}_{\mathrm{1}}^{( v )}$. The optimal solution of Eq. (\ref{E1}) is $\mathbb{S}_\frac{\alpha}{\mu}[\Theta^{( v )}]$, where $\mathbb{S}_\frac{\alpha}{\mu}[x]=sign(x)= \max(| x |-\frac{\alpha}{\mu}, 0)$ is the soft-thresholding operator~\cite{HaleYZ08}.

$\bullet$ \textbf{Solving $\bm{{\mathcal J}}$ with fixed ${\bf{C}}^{(v)}$, ${\bf{E}}^{(v)}$, ${\bf{P}}$ and ${\xi}^{(v)}$.} In this case, $\bm{\mathcal{J}}$ can be solved by
\begin{equation}\label{o4}
\begin{aligned}
{\bm{{\mathcal J}}^*} &= \mathop {\arg \min }\limits_{\bm{\mathcal {J}}} { \| \bm{{\mathcal J}} \|_{\Sp}^p} + \langle {\bm{\mathcal Y}_1},\bm{{\mathcal C} - \bm{\mathcal J}} \rangle  + \frac{\rho }{2}\| {\bm{{\mathcal C}} - \bm{{\mathcal J}}} \|_F^2\\
&= \mathop {\arg \min }\limits_{\bm{\mathcal J}} \frac{1}{\rho}{ \| \bm{{\mathcal J}} \|_{\Sp}^p} + \frac{1}{2}\| {\bm{{\mathcal C}}  + \frac{{\bm{\mathcal Y}_1}}{\rho }} - \bm{{\mathcal J}}\|_F^2
\end{aligned}
\end{equation}

To solve Eq. (\ref{o4}), we first introduce the Theorem~\ref{theorem1}~\cite{TPAMI}.
%
\begin{theorem}\label{theorem1}~\cite{TPAMI}
Suppose $\bm{{\mathcal Z}} \in {\mathbb{R}^{{n_1} \times {n_2} \times {n_3}}}$, $h = \min ({n_1},{n_2})$, let $\bm{{\mathcal Z}} = \bm{{\mathcal U}}*\bm{{\mathcal S}}*{\bm{{\mathcal V}}^T}$. For the following model:
\begin{equation}\label{t4-1}
\mathop {{\mathop{\rm argmin}\nolimits} }\limits_{\bm{\mathcal X}} \frac{1}{2}\| {\bm{\mathcal X}} - {\bm{{\mathcal Z}}} \|_F^2 + \tau {\| \bm{{\mathcal X}} \|_{\Sp}^p}
\end{equation}
the optimal solution ${{\bm{{\mathcal X}}}^*}$ is
\begin{equation}\label{t4-2}
{{\bm{{\mathcal X}}}^{\ast}} = {\Gamma _{\tau \cdot{n_3}}}({\bm{{\mathcal Z}}}) = {\bm{{\mathcal U}}}*ifft({{\bf{P}}_{\tau \cdot{n_3}}}(\overline {\bm{{\mathcal Z}}} ))*{{\bm{{\mathcal V}}}^\mathbf{T}}
\end{equation}
where, ${{\bf{P}}_{\tau \cdot{n_3}}}(\overline {\bm{\mathcal Z}} )$ is a tensor, ${{\bf{P}}_{\tau \cdot{n_3}}}({\overline {{\bm{\mathcal Z}}}^{(i)}})$ is the $i$-th frontal slice of ${{\bf{P}}_{\tau \cdot{n_3}}}(\overline {\bm{\mathcal Z}} )$.
\end{theorem}

According to Theorem~\ref{theorem1}, the solution of Eq. (\ref{o4}) is
\begin{equation}\label{o4-1}
{\bm{{\mathcal J}}^{\ast}}{\rm{ = }}{\Gamma _{\frac{1}{\rho }}}(\bm{{\mathcal C}} + \frac{1}{\rho }\bm{{\mathcal Y}}_1).
\end{equation}

$\bullet$ \textbf{Solving ${{\xi ^{(v)}}}$ with fixed other variables.} In this case, the optimization \wrt ${\xi ^{(v)}}$ in Eq. (\ref{o1}) is equivalent to
\begin{equation}\label{s-a-a}
\begin{aligned}
\mathop {\min }\limits_{\scriptstyle{\xi ^{(v)}}\hfill}\sum_{v=1}^V{\frac{tr(\mathbf{P}^{{\mathbf{T}}}\widetilde{\mathbf{L}}_{\mathbf{F}^{(v)}}\mathbf{P})}{{\xi ^{(v)}}}}, \quad \text {s.t.}\quad {\sum\limits_{v = 1}^V {{\xi ^{(v)}} = 1} ,{\xi ^{(v)}} \ge 0}
\end{aligned}
\end{equation}

Let $h^{(v)} = \sqrt{tr(\mathbf{P}^{{\mathbf{T}}}\widetilde{\mathbf{L}}_{\mathbf{F}^{(v)}}\mathbf{P})}$, the Eq. (\ref{s-a-a}) becomes
\begin{equation}\label{s-a}
\begin{aligned}
\mathop {\min }\limits_{\scriptstyle{\xi ^{(v)}}\hfill}\sum\limits_{v = 1}^V\frac{{h^{(v)}}^2}{{{\xi ^{(v)}}}}, \quad \text {s.t.}\quad {\sum\limits_{v = 1}^V {{\xi ^{(v)}} = 1} ,{\xi ^{(v)}} \ge 0}
\end{aligned}
\end{equation}

According to Cauchy-Schwartz's inequality, we have
\begin{equation}\label{s-a-c}
\begin{aligned}
\sum\limits_{v = 1}^V\frac{{h^{(v)}}^2}{{{\xi ^{(v)}}}}\overset{\left( \mathbf{i} \right)}{=}\left(\sum\limits_{v = 1}^V\frac{{h^{(v)}}^2}{{{\xi ^{(v)}}}}\right)\left(\sum_{v=1}^V{\xi^{(v)}}\right)\overset{\left( \mathbf{ii} \right)}{\geqslant}\left(\sum\limits_{v = 1}^V{h^{(v)}}\right)^2
\end{aligned}
\end{equation}
where ($\mathbf{i}$) holds because $\sum_{v=1}^V {{\xi ^{(v)}} = 1}$, ($\mathbf{ii}$) holds because $\sqrt{\xi ^{\left( v \right)}}\propto \frac{h^{\left( v \right)}}{\sqrt{\xi ^{\left( v \right)}}}$. The right-hand side in Eq. (\ref{s-a-c}) is a constant, therefore $\forall v=1, 2,\cdots, V$, the optimal ${\xi ^{(v)}}$ is
\begin{equation}\label{s-c}
\begin{aligned}
{\xi ^{(v)}}^{+} := h^{(v)}/{{\sum_{v = 1}^V h^{(v)}}}
\end{aligned}
\end{equation}

Finally, the optimization procedure for solving the model (\ref{o1}) is outlined in Algorithm~\ref{A1}.

In this paper, we utilize the same way as reported in~\cite{SFMC} to construct graphs $\mathbf{B}^{(v)} \in \mathbb{R}^{n \times m}$, $v=1, 2, \cdots, V$.
\begin{table*}[!h]
\begin{center}
\caption{Summary of computational complexity, where $V, m, n$ and $K$ are the number of views, anchors, data points and clusters, respectively. $t$ is the iteration number. $m \ll n$, N/A means not applicable.}\label{2com}
\vspace{-4mm}
\resizebox{2.0\columnwidth}{!}{
\begin{tabular}{c!{\vrule width1.0pt}c!{\vrule width1.0pt}c!{\vrule width1.0pt}c!{\vrule width1.0pt}c!{\vrule width1.0pt}c!{\vrule width1.0pt}c}
\Xhline{1.5pt}
Method&Construction of Graphs&Solving $\bm{\mathbf{C}}^{(v)}$&Solving $\bm{\mathcal{J}}$&Solving $\bm{\mathbf{E}}^{(v)}$&Solving $\bm{\mathbf{P}}$&Total\\
\hline
ETLMSC&$\bm{\mathcal{O}}(Vn^2d)$&N/A&$\bm{\mathcal{O}}(Vn^2\log(Vn)+V^2n^2)$ &$\bm{\mathcal{O}}(Vn^2)$ &$\bm{\mathcal{O}}(n^3)$&$\bm{\mathcal{O}}(n^3+Vn^2d+Vn^2t\log(Vn))$\\
\hline
Ours&$\bm{\mathcal{O}}(Vnmd + Vnm\log(m))$&$\bm{\mathcal{O}}(VnmK+Vnm\log(m))$&$\bm{\mathcal{O}}(Vnm\log(Vn)+V^2mn)$&$\bm{\mathcal{O}}(Vnm)$&$\bm{\mathcal{O}}(Vnm + m^2n)$&$\bm{\mathcal{O}}(Vnmd+m^2nt)$\\
\Xhline{1.5pt}
\end{tabular}}
\end{center}
\end{table*}
\subsection{Complexity Analysis}
Our method consists of two stages: 1) Construction of graphs $\{\mathbf{B}^{(v)}\}_{v=1}^V$, same to~\cite{SFMC}, 2) optimization by iteratively solving Eq. (\ref{o1}). The first stage takes $\bm{\mathcal{O}}(Vnmd + Vnm\log(m))$ time, where $V$, $m$ and $n$ are the number of views, anchors and samples, respectively. The second stage mainly focuses on four variables ($\bm{\mathbf{C}}^{(v)}$, $\bm{\mathcal{J}}$, $\bm{\mathbf{E}}^{(v)}$, $\bm{\mathbf{P}}$), the complexity in updating these variables iteratively are $\bm{\mathcal{O}}(VnmK+Vnm\log(m))$, $\bm{\mathcal{O}}(Vnm\log(Vn)+V^2mn)$, $\bm{\mathcal{O}}(Vnm)$ and $\bm{\mathcal{O}}(Vnm + m^2n)$, where $K$ and $t$ are the number of clusters and iteration, respectively. Due to $m\ll n$, the main complexity in this stage is $\bm{\mathcal{O}}(m^2nt + Vnmt\log(Vn))$. Therefore, the main computational complexity of our method is actually $\bm{\mathcal{O}}(m^2nt + Vnmd)$, which is linear to the $n$. The main computational complexity of ETLMSC and our method are summarized in Table~\ref{2com}.
\vspace{-2mm}
\begin{algorithm}[!t]
  \caption{Effective and Efficient Graph Learning for Multi-view Clustering}
  \label{A1}
  \LinesNumbered
  \KwIn{Data matrices: $\{\mathbf{X}^{(v)}\}_{v=1}^V\in \mathbb{R}^{n \times d_v}$, anchors number $m$, and cluster number $K$, $\alpha$.}
  \KwOut{Graph $\bm{\mathbf{C}}$ with $K$-connected components.}
  Construct graphs $\mathbf{B}^{(v)} \in \mathbb{R}^{n \times m}$ like~\cite{SFMC}\;
  Initialize $\mathbf{C}^{(v)}=\mathbf{B}^{(v)}$, $\mathbf{E}^{(v)}=0$, $\mathbf{Y}_1^{(v)}=0$, $\bm{\mathcal{Y}}_2 = 0$, $\bm{\mathcal{J}} = 0$,
  $\rho=10^{-5}$, $\mu=10^{-5}$, $\eta=1.1$, $\xi^{(v)}=1/V$\;
  \While{not converge}
  {
  Update $\mathbf{P}$ by solving Eq. (\ref{P-1})\;
  Update $\{\mathbf{C}^{(v)}\}_{v=1}^V$ by solving Eq. (\ref{t9})\;
  Update $\{\mathbf{E}^{(v)}\}_{v=1}^V$ by solving Eq. (\ref{E1})\;
  Update $\bm{\mathcal{J}}$ by using Eq. (\ref{o4-1})\;
  Update $\xi^{(v)}$ by using Eq. (\ref{s-c})\;
  Update $\mathbf{Y}_1^{(v)}$, $\bm{\mathcal{Y}}_2$, $\mu$ and $\rho$: $\mathbf{Y}_1^{(v)}:=\mathbf{Y}_1^{(v)}+\mu(\mathbf{B}^{(v)}-\mathbf{C}^{(v)}-\mathbf{E}^{(v)})$, $\bm{\mathcal{Y}}_2 := \bm{\mathcal{Y}}_2 + \rho(\bm{\mathcal{C}}-\bm{\mathcal{J}})$, $\mu := \eta \mu$, $\rho := \eta \rho$\;
  }
  Directly achieve the $K$ clusters based on the connectivity of $\mathbf{C}={\sum_{v=1}^V \frac{\mathbf{C}^{(v)}}{\xi^{(v)}}}/{\sum_{v=1}^V \frac{1}{\xi^{(v)}}}$\;
  \textbf{return:} Clustering results.
\end{algorithm}
\section{Experiments}
\subsection{Experimental Setup}
\textbf{Datasets:} We use the following multi-view datasets to investigate the superiority of our proposed method: (1) \emph{\textbf{MSRC-v5}}~\cite{WinnJ05} includes $7$ kinds of objects with $210$ images. Same to~\cite{SFMC}, we choose $24$-dimension (D) CM feature, $576$-D HOG feature, $512$-D GIST feature, $256$-D LBP feature, $254$-D CENT feature as $5$ views. (2) \emph{\textbf{Handwritten4}}~\cite{Dua-2019} includes $10$ digits with $2,000$ images generated from UCI machine learning repository. $76$-D FOU feature, $216$-D FAC feature, $47$-D ZER feature and $6$-D MOR feature are employed as $4$ views. (3) \emph{\textbf{Mnist4}}~\cite{Deng12} includes $4$ categories handwritten digits, \ie, from digit $0$ to digit $3$, with $4,000$ images. We utilize $30$-D ISO feature, $9$-D LDA feature and $30$-D NPE feature as $3$ views. (4) \emph{\textbf{Caltech101-20}}~\cite{Fei-FeiFP07} includes $20$ categories with $2,386$ images. It is a subsets of Caltech101 datasets. We employ $48$-D GABOR feature, $40$-D WM feature, $254$-D CENT feature, $1,984$-D HOG feature, $512$-D GIST feature and $928$-D LBP feature as $6$ views. (5) \emph{\textbf{NUS-WIDE}}~\cite{ChuaTHLLZ09} has $31$ categories with $30,000$ object images. $64$-D CH feature, $225$-D CM feature, $144$-D CORR feature, $73$-D EDH feature and $128$-D WT feature are adopted as $5$ views. (6) \emph{\textbf{Reuters}}~\cite{ApteDW94} has $18,758$ documents of $6$ categories. We adopt $21,513$-D English, $24,892$-D France, $34,251$-D German, $15,506$-D Italian and $11,547$-D Spanish as $5$ views.

\textbf{Baselines:} We compare our method with $13$ clustering methods: single-view constrained Laplacian rank (s-CLR)~\cite{NieWJH16}, Co-train~\cite{KumarD11}, Co-reg~\cite{KumarRD11r}, SwMC~\cite{NieLL17}, MVGL~\cite{ZhanZGW18}, MVSC~\cite{LiNHH15}, RDEKM~\cite{XuHNL17}, SMSC~\cite{HuNWL20}, AMGL~\cite{NieLL16}; MLAN~\cite{NieCLL18}, large-scale multi-view subspace clustering (LMVSC)~\cite{KangZZSHX20}, ETLMSC~\cite{WuLZ19} and SFMC~\cite{SFMC}.

\textbf{Metrics:} The widely used $7$ metrics are applied to evaluate the clustering performance, \ie, 1) Accuracy (ACC); 2) Normalized Mutual Information (NMI); 3) Purity; 4) Precision (PRE); 5) Recall (REC); 6) F-score; and 7) Adjusted Rand Index (ARI). For all metrics, the higher value indicates the better clustering performance. For more detailed definitions about each of the metrics, please refer to~\cite{xie2018hyper}.
\begin{table*}[!t]
\begin{center}
\caption{The clustering performances on MSRC-v5 and Handwritten4 datasets.}
\label{result1}
\vspace{-2mm}
\resizebox{1.5\columnwidth}{!}{
\begin{tabular}{c!{\vrule width1.2pt}ccccccc}
\Xhline{2pt}
Dataset&\multicolumn{7}{c}{\textbf{MSRC-v5}}\\
\hline
Metric&ACC&NMI&Purity&PER&REC&F-score&ARI\\
\Xhline{1.2pt}
s-CLR ($\mathbf{X}^{(1)}$)~\cite{NieWJH16}&0.333$\pm$0.000&0.226$\pm$0.000&0.381$\pm$0.000&0.199$\pm$0.000&0.507$\pm$0.000&0.286$\pm$0.000&0.108$\pm$0.000\\
s-CLR ($\mathbf{X}^{(2)}$)~\cite{NieWJH16}&0.681$\pm$0.000&0.608$\pm$0.000&0.710$\pm$0.000&0.478$\pm$0.000&0.707$\pm$0.000&0.570$\pm$0.000&0.471$\pm$0.000\\
s-CLR ($\mathbf{X}^{(3)}$)~\cite{NieWJH16}&0.648$\pm$0.000&0.595$\pm$0.000&0.652$\pm$0.000&0.428$\pm$0.000&0.688$\pm$0.000&0.528$\pm$0.000&0.467$\pm$0.000\\
s-CLR ($\mathbf{X}^{(4)}$)~\cite{NieWJH16}&0.400$\pm$0.000&0.397$\pm$0.000&0.486$\pm$0.000&0.307$\pm$0.000&0.421$\pm$0.000&0.355$\pm$0.000&0.231$\pm$0.000\\
s-CLR ($\mathbf{X}^{(5)}$)~\cite{NieWJH16}&0.614$\pm$0.000&0.529$\pm$0.000&0.638$\pm$0.000&0.461$\pm$0.000&0.634$\pm$0.000&0.534$\pm$0.000&0.445$\pm$0.000\\
s-CLR-Concat~\cite{NieWJH16}&0.590$\pm$0.000&0.509$\pm$0.000&0.614$\pm$0.000&0.451$\pm$0.000&0.482$\pm$0.000&0.466$\pm$0.000&0.377$\pm$0.000\\
Co-train~\cite{KumarD11}&0.674$\pm$0.070&0.587$\pm$0.046&0.690$\pm$0.061&0.546$\pm$0.064&0.571$\pm$0.063&0.558$\pm$0.063&0.485$\pm$0.074\\
Co-reg~\cite{KumarRD11r}&0.635$\pm$0.007&0.578$\pm$0.006&0.659$\pm$0.006&0.511$\pm$0.008&0.535$\pm$0.007&0.522$\pm$0.007&0.425$\pm$0.030\\
SwMC~\cite{NieLL17}&0.776$\pm$0.000&0.774$\pm$0.000&0.805$\pm$0.000&0.687$\pm$0.000&0.831$\pm$0.000&0.752$\pm$0.000&0.708$\pm$0.000\\
MVGL~\cite{ZhanZGW18}&0.690$\pm$0.000&0.663$\pm$0.000&0.733$\pm$0.000&0.466$\pm$0.000&0.715$\pm$0.000&0.564$\pm$0.000&0.476$\pm$0.000\\
MVSC~\cite{LiNHH15}&0.794$\pm$0.075&0.672$\pm$0.058&0.756$\pm$0.071&0.585$\pm$0.091&0.779$\pm$0.035&0.664$\pm$0.062&0.600$\pm$0.079\\
RDEKM~\cite{XuHNL17}&0.738$\pm$0.000&0.650$\pm$0.000&0.738$\pm$0.000&0.594$\pm$0.000&0.647$\pm$0.000&0.619$\pm$0.000&0.555$\pm$0.000\\
SMSC~\cite{HuNWL20}&0.766$\pm$0.000&0.717$\pm$0.000&0.804$\pm$0.000&0.672$\pm$0.000&0.718$\pm$0.000&0.694$\pm$0.000&0.643$\pm$0.000\\
AMGL~\cite{NieLL16}&0.751$\pm$0.078&0.704$\pm$0.044&0.789$\pm$0.056&0.621$\pm$0.090&0.744$\pm$0.026&0.674$\pm$0.063&0.615$\pm$0.079\\
MLAN~\cite{NieCLL18}&0.681$\pm$0.000&0.630$\pm$0.000&0.733$\pm$0.000&0.494$\pm$0.000&0.718$\pm$0.000&0.694$\pm$0.000&0.643$\pm$0.000\\
ETLMSC~\cite{WuLZ19}&0.962$\pm$0.000&0.937$\pm$0.000&0.962$\pm$0.000&0.926$\pm$0.000&0.931$\pm$0.000&0.928$\pm$0.000&0.917$\pm$0.000\\
SFMC~\cite{SFMC}&0.810$\pm$0.000&0.721$\pm$0.000&0.810$\pm$0.000&0.657$\pm$0.000&0.782$\pm$0.000&0.714$\pm$0.000&0.663$\pm$0.000\\
\textbf{Ours}&\textbf{0.995}$\pm$\textbf{0.000}&\textbf{0.989}$\pm$\textbf{0.000}&\textbf{0.995}$\pm$\textbf{0.000}&\textbf{0.990}$\pm$\textbf{0.000}&\textbf{0.990}$\pm$\textbf{0.000}&\textbf{0.990}$\pm$\textbf{0.000}&\textbf{0.989}$\pm$\textbf{0.000}\\
\Xhline{2pt}
Dataset&\multicolumn{7}{c}{\textbf{Handwritten4}}\\
\hline
Metric&ACC&NMI&Purity&PER&REC&F-score&ARI\\
\Xhline{1.2pt}
s-CLR ($\mathbf{X}^{(1)}$)~\cite{NieWJH16}&0.660$\pm$0.000&0.683$\pm$0.000&0.699$\pm$0.000&0.527$\pm$0.000&0.722$\pm$0.000&0.609$\pm$0.000&0.558$\pm$0.000\\
s-CLR ($\mathbf{X}^{(2)}$)~\cite{NieWJH16}&0.698$\pm$0.000&0.731$\pm$0.000&0.731$\pm$0.000&0.592$\pm$0.000&0.803$\pm$0.000&0.681$\pm$0.000&0.640$\pm$0.000\\
s-CLR ($\mathbf{X}^{(3)}$)~\cite{NieWJH16}&0.660$\pm$0.000&0.651$\pm$0.000&0.661$\pm$0.000&0.441$\pm$0.000&0.760$\pm$0.000&0.558$\pm$0.000&0.495$\pm$0.000\\
s-CLR ($\mathbf{X}^{(4)}$)~\cite{NieWJH16}&0.403$\pm$0.000&0.452$\pm$0.000&0.426$\pm$0.000&0.310$\pm$0.000&0.383$\pm$0.000&0.343$\pm$0.000&0.262$\pm$0.000\\
s-CLR-Concat~\cite{NieWJH16}&0.759$\pm$0.000&0.751$\pm$0.000&0.760$\pm$0.000&0.610$\pm$0.000&0.865$\pm$0.000&0.716$\pm$0.000&0.678$\pm$0.000\\
Co-train~\cite{KumarD11}&0.381$\pm$0.021&0.301$\pm$0.019&0.399$\pm$0.019&0.290$\pm$0.018&0.295$\pm$0.019&0.293$\pm$0.018&0.214$\pm$0.021\\
Co-reg~\cite{KumarRD11r}&0.784$\pm$0.010&0.758$\pm$0.004&0.795$\pm$0.008&0.698$\pm$0.010&0.724$\pm$0.005&0.710$\pm$0.007&0.667$\pm$0.037\\
SwMC~\cite{NieLL17}&0.758$\pm$0.000&0.833$\pm$0.000&0.792$\pm$0.000&0.686$\pm$0.000&0.867$\pm$0.000&0.766$\pm$0.000&0.737$\pm$0.000\\
MVGL~\cite{ZhanZGW18}&0.811$\pm$0.000&0.809$\pm$0.000&0.831$\pm$0.000&0.721$\pm$0.000&0.826$\pm$0.000&0.770$\pm$0.000&0.743$\pm$0.000\\
MVSC~\cite{LiNHH15}&0.796$\pm$0.059&0.820$\pm$0.030&0.808$\pm$0.044&0.715$\pm$0.082&0.838$\pm$0.035&0.769$\pm$0.046&0.741$\pm$0.053\\
RDEKM~\cite{XuHNL17}&0.805$\pm$0.000&0.803$\pm$0.000&0.842$\pm$0.000&0.714$\pm$0.000&0.806$\pm$0.000&0.757$\pm$0.000&0.728$\pm$0.000\\
SMSC~\cite{HuNWL20}&0.742$\pm$0.000&0.781$\pm$0.000&0.759$\pm$0.000&0.675$\pm$0.000&0.767$\pm$0.000&0.717$\pm$0.000&0.685$\pm$0.000\\
AMGL~\cite{NieLL16}&0.704$\pm$0.045&0.762$\pm$0.040&0.732$\pm$0.042&0.591$\pm$0.081&0.781$\pm$0.022&0.670$\pm$0.060&0.628$\pm$0.070\\
MLAN~\cite{NieCLL18}&0.778$\pm$0.045&0.832$\pm$0.027&0.812$\pm$0.045&0.706$\pm$0.053&0.871$\pm$0.017&0.779$\pm$0.039&0.752$\pm$0.044\\
ETLMSC~\cite{WuLZ19}&0.938$\pm$0.001&0.893$\pm$0.001&0.938$\pm$0.001&0.886$\pm$0.001&0.890$\pm$0.001&0.888$\pm$0.001&0.876$\pm$0.001\\
SFMC~\cite{SFMC}&0.853$\pm$0.000&0.871$\pm$0.000&0.873$\pm$0.000&0.775$\pm$0.000&0.910$\pm$0.000&0.837$\pm$0.000&0.817$\pm$0.000\\
\textbf{Ours}&\textbf{0.995}$\pm$\textbf{0.000}&\textbf{0.986}$\pm$\textbf{0.000}&\textbf{0.995}$\pm$\textbf{0.000}&\textbf{0.989}$\pm$\textbf{0.000}&\textbf{0.989}$\pm$\textbf{0.000}&\textbf{0.989}$\pm$\textbf{0.000}&\textbf{0.988}$\pm$\textbf{0.000}\\
\Xhline{2pt}
\end{tabular}}
\end{center}
\vspace{-4mm}
\end{table*}

\begin{table*}[!t]
\begin{center}
\caption{The clustering performances on Mnist4 and Caltech101-20 datasets.}
\label{result2}
\vspace{-2mm}
\resizebox{1.5\columnwidth}{!}{
\begin{tabular}{c!{\vrule width1.2pt}ccccccc}
\Xhline{2pt}
Dataset&\multicolumn{7}{c}{\textbf{Mnist4}}\\
\hline
Metric&ACC&NMI&Purity&PER&REC&F-score&ARI\\
\Xhline{1.2pt}
s-CLR ($\mathbf{X}^{(1)}$)~\cite{NieWJH16}&0.660$\pm$0.000&0.679$\pm$0.000&0.742$\pm$0.000&0.626$\pm$0.000&0.798$\pm$0.000&0.701$\pm$0.000&0.585$\pm$0.000\\
s-CLR ($\mathbf{X}^{(2)}$)~\cite{NieWJH16}&0.843$\pm$0.000&0.762$\pm$0.000&0.744$\pm$0.000&0.640$\pm$0.000&0.824$\pm$0.000&0.721$\pm$0.000&0.655$\pm$0.000\\
s-CLR ($\mathbf{X}^{(3)}$)~\cite{NieWJH16}&0.743$\pm$0.000&0.661$\pm$0.000&0.744$\pm$0.000&0.642$\pm$0.000&0.827$\pm$0.000&0.723$\pm$0.000&0.657$\pm$0.000\\
s-CLR-Concat$^\dagger$~\cite{NieWJH16}&0.897$\pm$0.000&0.747$\pm$0.000&0.897$\pm$0.000&0.813$\pm$0.000&0.822$\pm$0.000&0.817$\pm$0.000&0.756$\pm$0.000\\
Co-train~\cite{KumarD11}&0.758$\pm$0.001&0.554$\pm$0.001&0.758$\pm$0.001&0.643$\pm$0.001&0.644$\pm$0.001&0.639$\pm$0.001&0.518$\pm$0.002\\
Co-reg~\cite{KumarRD11r}&0.785$\pm$0.003&0.602$\pm$0.001&0.786$\pm$0.002&0.670$\pm$0.002&0.696$\pm$0.002&0.682$\pm$0.001&0.575$\pm$0.002\\
SwMC~\cite{NieLL17}&0.914$\pm$0.000&0.799$\pm$0.000&0.912$\pm$0.000&0.844$\pm$0.000&0.852$\pm$0.000&0.848$\pm$0.000&0.799$\pm$0.000\\
MVGL~\cite{ZhanZGW18}&0.912$\pm$0.000&0.785$\pm$0.000&0.910$\pm$0.000&0.795$\pm$0.000&0.804$\pm$0.000&0.800$\pm$0.000&0.733$\pm$0.000\\
MVSC~\cite{LiNHH15}&0.733$\pm$0.115&0.651$\pm$0.069&0.780$\pm$0.070&0.650$\pm$0.092&0.773$\pm$0.041&0.704$\pm$0.066&0.592$\pm$0.096\\
RDEKM~\cite{XuHNL17}&0.885$\pm$0.000&0.717$\pm$0.000&0.885$\pm$0.000&0.795$\pm$0.000&0.804$\pm$0.000&0.800$\pm$0.000&0.733$\pm$0.000\\
SMSC~\cite{HuNWL20}&0.913$\pm$0.000&0.789$\pm$0.000&0.913$\pm$0.000&0.843$\pm$0.000&0.850$\pm$0.000&0.846$\pm$0.000&0.795$\pm$0.000\\
AMGL~\cite{NieLL16}&0.910$\pm$0.000&0.785$\pm$0.000&0.910$\pm$0.000&0.836$\pm$0.000&0.843$\pm$0.000&0.840$\pm$0.000&0.786$\pm$0.000\\
MLAN~\cite{NieCLL18}&0.744$\pm$0.001&0.659$\pm$0.001&0.744$\pm$0.000&0.643$\pm$0.001&\textbf{0.921}$\pm$\textbf{0.001}&0.757$\pm$0.001&0.656$\pm$0.001\\
ETLMSC~\cite{WuLZ19}&0.934$\pm$0.000&0.847$\pm$0.000&0.934$\pm$0.000&0.878$\pm$0.000&0.885$\pm$0.000&0.881$\pm$0.000&0.842$\pm$0.000\\
SFMC~\cite{SFMC}&0.917$\pm$0.000&0.801$\pm$0.000&0.917$\pm$0.000&0.846$\pm$0.000&0.855$\pm$0.000&0.852$\pm$0.000&0.802$\pm$0.000\\
\textbf{Ours}&\textbf{0.938}$\pm$\textbf{0.000}&\textbf{0.855}$\pm$\textbf{0.000}&\textbf{0.938}$\pm$\textbf{0.000}&\textbf{0.885}$\pm$\textbf{0.000}&0.890$\pm$0.000&\textbf{0.888}$\pm$\textbf{0.000}&\textbf{0.850}$\pm$\textbf{0.000}\\
\Xhline{2pt}
Dataset&\multicolumn{7}{c}{\textbf{Caltech101-20}}\\
\hline
Metric&ACC&NMI&Purity&PER&REC&F-score&ARI\\
\Xhline{1.2pt}
s-CLR ($\mathbf{X}^{(1)}$)~\cite{NieWJH16}&0.390$\pm$0.000&0.174$\pm$0.000&0.450$\pm$0.000&0.195$\pm$0.000&0.720$\pm$0.000&0.307$\pm$0.000&0.069$\pm$0.000\\
s-CLR ($\mathbf{X}^{(2)}$)~\cite{NieWJH16}&0.387$\pm$0.000&0.238$\pm$0.000&0.468$\pm$0.000&0.177$\pm$0.000&0.501$\pm$0.000&0.261$\pm$0.000&0.027$\pm$0.000\\
s-CLR ($\mathbf{X}^{(3)}$)~\cite{NieWJH16}&0.323$\pm$0.000&0.206$\pm$0.000&0.422$\pm$0.000&0.169$\pm$0.000&0.515$\pm$0.000&0.254$\pm$0.000&0.013$\pm$0.000\\
s-CLR ($\mathbf{X}^{(4)}$)~\cite{NieWJH16}&0.442$\pm$0.000&0.269$\pm$0.000&0.492$\pm$0.000&0.198$\pm$0.000&0.745$\pm$0.000&0.313$\pm$0.000&0.076$\pm$0.000\\
s-CLR ($\mathbf{X}^{(5)}$)~\cite{NieWJH16}&0.414$\pm$0.000&0.284$\pm$0.000&0.479$\pm$0.000&0.205$\pm$0.000&0.763$\pm$0.000&0.324$\pm$0.000&0.091$\pm$0.000\\
s-CLR ($\mathbf{X}^{(6)}$)~\cite{NieWJH16}&0.358$\pm$0.000&0.244$\pm$0.000&0.452$\pm$0.000&0.172$\pm$0.000&0.611$\pm$0.000&0.268$\pm$0.000&0.019$\pm$0.000\\
s-CLR-Concat~\cite{NieWJH16}&0.596$\pm$0.000&0.429$\pm$0.000&0.653$\pm$0.000&0.313$\pm$0.000&\textbf{0.817}$\pm$\textbf{0.000}&0.453$\pm$0.000&0.285$\pm$0.000\\
Co-train~\cite{KumarD11}&0.397$\pm$0.020&0.510$\pm$0.016&0.737$\pm$0.015&0.690$\pm$0.034&0.233$\pm$0.015&0.349$\pm$0.020&0.290$\pm$0.022\\
Co-reg~\cite{KumarRD11r}&0.412$\pm$0.006&0.587$\pm$0.003&0.754$\pm$0.004&0.712$\pm$0.008&0.243$\pm$0.004&0.363$\pm$0.006&0.295$\pm$0.025\\
SwMC~\cite{NieLL17}&0.599$\pm$0.000&0.493$\pm$0.000&0.700$\pm$0.000&0.509$\pm$0.000&0.625$\pm$0.000&0.431$\pm$0.000&0.265$\pm$0.000\\
MVGL~\cite{ZhanZGW18}&0.600$\pm$0.000&0.474$\pm$0.000&0.696$\pm$0.000&0.325$\pm$0.000&0.653$\pm$0.000&0.440$\pm$0.000&0.282$\pm$0.000\\
MVSC~\cite{LiNHH15}&0.595$\pm$0.000&0.613$\pm$0.000&0.717$\pm$0.000&0.542$\pm$0.000&0.546$\pm$0.000&0.541$\pm$0.000&0.451$\pm$0.000\\
RDEKM~\cite{XuHNL17}&0.424$\pm$0.000&0.572$\pm$0.000&0.768$\pm$0.000&0.747$\pm$0.000&0.299$\pm$0.000&0.427$\pm$0.000&0.368$\pm$0.000\\
SMSC~\cite{HuNWL20}&0.582$\pm$0.000&0.590$\pm$0.000&0.748$\pm$0.000&0.701$\pm$0.000&0.473$\pm$0.000&0.565$\pm$0.000&0.485$\pm$0.000\\
AMGL~\cite{NieLL16}&0.557$\pm$0.047&0.552$\pm$0.061&0.677$\pm$0.058&0.480$\pm$0.093&0.539$\pm$0.015&0.503$\pm$0.054&0.397$\pm$0.080\\
MLAN~\cite{NieCLL18}&0.526$\pm$0.007&0.474$\pm$0.003&0.666$\pm$0.000&0.279$\pm$0.003&0.559$\pm$0.020&0.372$\pm$0.007&0.198$\pm$0.007\\
ETLMSC~\cite{WuLZ19}&0.483$\pm$0.017&0.681$\pm$0.007&0.845$\pm$0.013&\textbf{0.832}$\pm$\textbf{0.017}&0.275$\pm$0.007&0.413$\pm$0.010&0.362$\pm$0.010\\
SFMC~\cite{SFMC}&0.642$\pm$0.000&0.595$\pm$0.000&0.748$\pm$0.000&0.586$\pm$0.000&0.677$\pm$0.000&0.628$\pm$0.000&0.461$\pm$0.000\\
\textbf{Ours}&\textbf{0.781}$\pm$\textbf{0.000}&\textbf{0.791}$\pm$\textbf{0.000}&\textbf{0.866}$\pm$\textbf{0.000}&0.672$\pm$0.000&0.696$\pm$0.000&\textbf{0.684}$\pm$\textbf{0.000}&\textbf{0.621}$\pm$\textbf{0.000}\\
\Xhline{2pt}
\end{tabular}}
\vspace{-4mm}
\end{center}
\end{table*}
\vspace{-2mm}

\subsection{Comparisons with State-of-the-art Methods}

Tables~\ref{result1},~\ref{result2},~\ref{result3} present the metrics comparison of the above methods on six datasets. For each experiment, we independently repeat the involved methods $20$ times and show the averages with corresponding standard deviations. For CLR with single-view setting, s-CLR ($\mathbf{X}^{(i)}$) denotes the results of CLR by employing features in $i$-th view, and s-CLR-Concat denotes the results of s-CLR on the concatenated view-features. From Tables~\ref{result1},~\ref{result2},~\ref{result3}, we have the following interesting observations:

\begin{itemize}
  \item Single-view clustering method s-CLR is overall inferior to multi-view clustering methods. This is probably because that information embedded in different views are complementary and multi-view methods well use this formation which is important for improving clustering performance, while s-CLR does not. Moreover, the performance of s-CLR has a significant difference on different views. The reason may be that each view contains some content of the object that other views do not. Thus, graphs, which are constructed from different views, usually has significant different roles for clustering.
  \item Multi-view clustering methods Co-train and Co-reg are overall inferior to the other multi-view methods. This is probably because that Co-train and Co-reg neglect the significant difference between different views for clustering. Another reason may be that their performance heavily depend on the graphs which are artificially defined. However, in real real-world applications, it is difficult to artificially select a suitable graph for complex data.
  \item Our proposed method and ETLMSC are superior to the other methods. The reason may be that our method and ETLMSC well exploit the complementary information and high-order information embedded in graphs of different views, while other methods do not.
  \item Our method is remarkably superior to SFMC. For example, on MSRC-v5 dataset, compared with SFMC, our method gains significant improvement around $18.5\%$, $26.8\%$, $18.5\%$, $33.3\%$, $20.8\%$, $27.6\%$, and $32.6\%$ in terms of ACC, NMI, Purity, PER, REC, F-score, and ARI, respectively.  For multi-view clustering, an ideal view-similar graph should have both the low-rank structure and K-connected components. To get the best clustering performance, view-similar graphes between different views have not only the high similarity but also high-similar spatial geometric structure. Our method explicitly takes into account these important information by minimization of tensor Schatten $p$-norm, while SFMC does not. Moreover, in our method, rank of the learned graph approximates the target rank, while SFMC does not.
  \item Although our method is an anchor-based method, the performances of our method also superior to ETLMSC. The reason may be that our method explicitly takes into account the significant difference of different views. Moreover, our method  well characterizes the cluster structure and does not require any post-processing to final clustering results. This indicates that anchors totally well encode cluster structure of data and afford efficient clustering.
  \item Towards large scale datasets, due to CPU limitations, some methods, \eg, co-train, SwMC, MVGL, MLAN, SMSC, AMGL and ETLMSC, suffer from the out-of-memory issue. Thus, we herein compare the clustering performances of our method with partial competitors in Table~\ref{result3}, where the number of anchors is set to $37$ and $400$ on Reuters dataset and NUS-WID dataset, respectively. As shown in Table~\ref{result3}, the proposed method significantly and consistently outperforms all competitors, indicating the effectiveness of large scale multi-view data clustering.
\end{itemize}
\begin{table}[!t]
\begin{center}
\caption{The results on large-scale datasets, ``OM'' is ``out-of-memory error''.}\label{result3}
\vspace{-4mm}
\resizebox{1.0\columnwidth}{!}{
\begin{tabular}{c!{\vrule width1.0pt}ccc!{\vrule width1.0pt}ccc}
\Xhline{1.5pt}
Dataset&\multicolumn{3}{c!{\vrule width1.0pt}}{\textbf{NUS-WIDE}}&\multicolumn{3}{c}{\textbf{Reuters}}\\
\hline
Metric&ACC&NMI&Purity&ACC&NMI&Purity\\
\Xhline{1.2pt}
Co-reg~\cite{KumarRD11r}&0.1194&0.1143&0.2144&0.5627&0.3261&0.5523\\
MVSC~\cite{LiNHH15}&0.1496&0.0752&0.1839&0.5958&0.3472&0.5741\\
LMVSC~\cite{KangZZSHX20}&0.1140&0.0768&0.1708&0.5890&0.3346&0.6145\\
ETLMSC~\cite{WuLZ19}&OM&OM&OM&OM&OM&OM\\
SFMC~\cite{SFMC}&0.1689&0.0601&0.1904&0.6023&0.3541&0.6042\\
\textbf{Ours}&\textbf{0.2163}&\textbf{0.1475}&\textbf{0.2187}&\textbf{0.6512}&\textbf{0.4886}&\textbf{0.7020}\\
\Xhline{1.5pt}
\end{tabular}}
\vspace{-4mm}
\end{center}
\end{table}

\begin{figure}[!t]
\centering
\subfigure[MSRC-v5]{
\begin{minipage}[t]{0.46\linewidth}
\centering
\includegraphics[width=4.2cm]{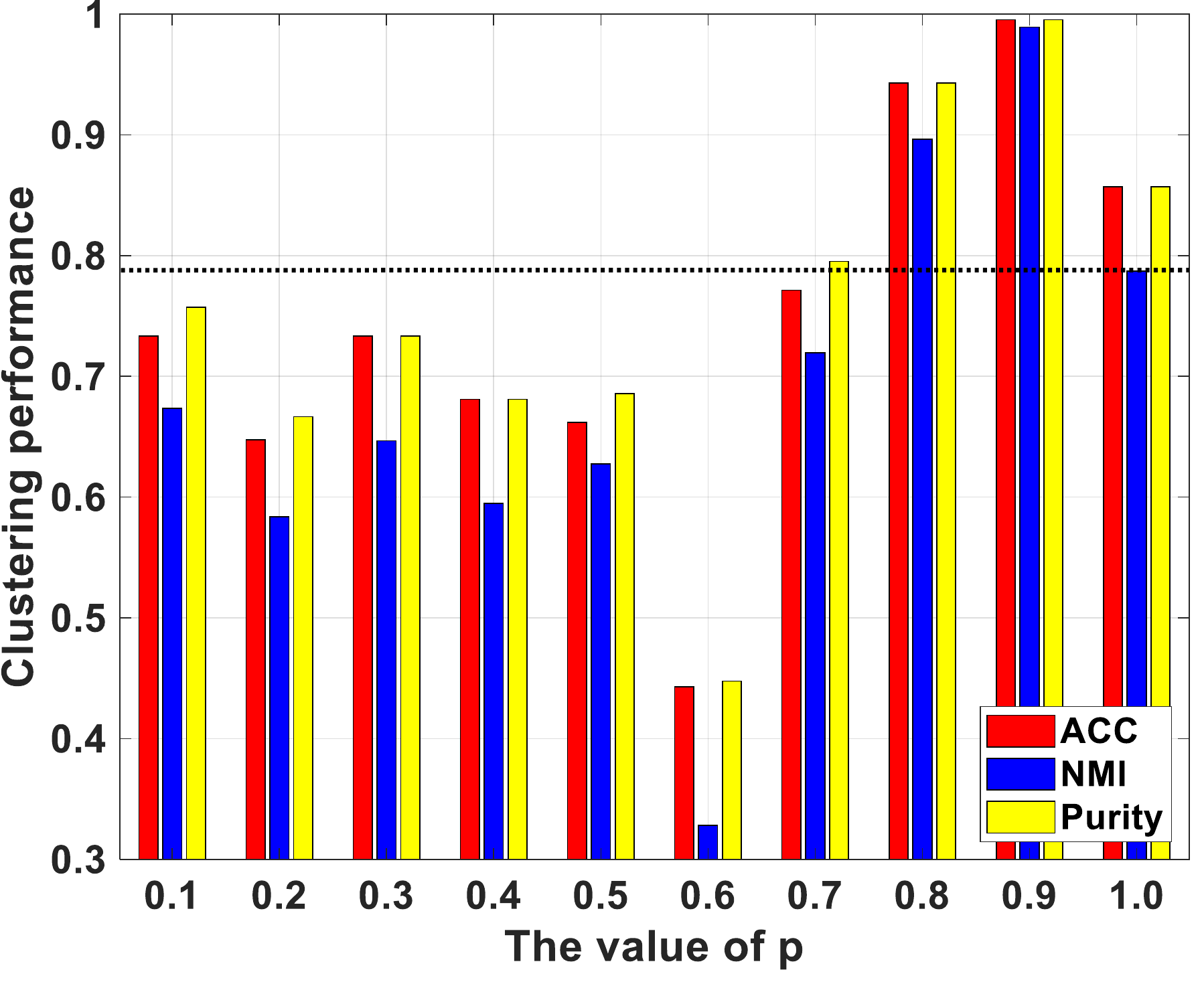}
\end{minipage}
}
\subfigure[Handwritten4]{
\begin{minipage}[t]{0.48\linewidth}
\centering
\includegraphics[width=4.2cm]{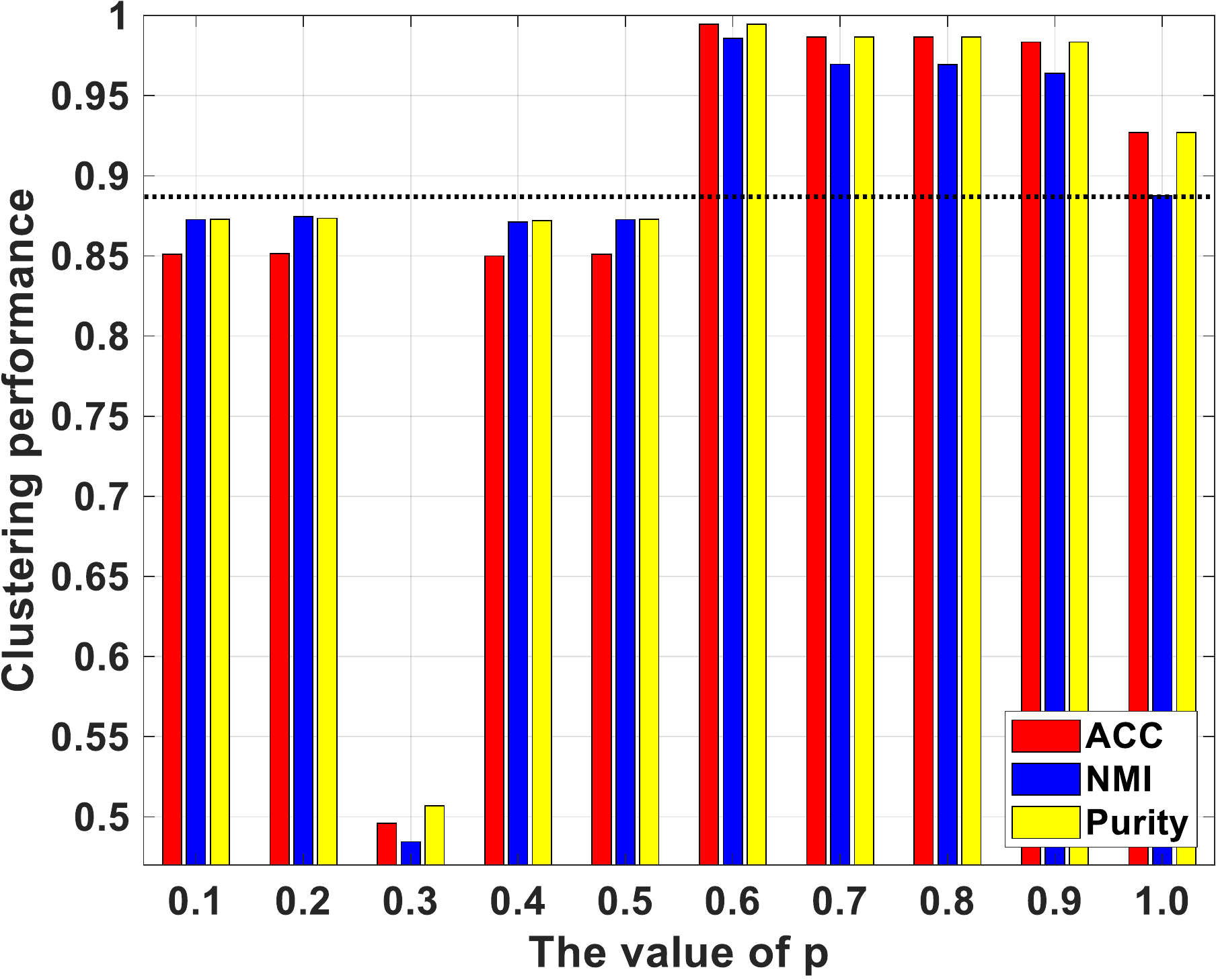}
\end{minipage}
}
\centering
\caption{The clustering performances of our method with the varying value of $p$ on MSRC-v5 and Handwritten4 datasets.}
\label{p-value}
\vspace{-4mm}
\end{figure}
\begin{figure}[!t]
\centering
\subfigure[MSRC-v5]{
\begin{minipage}[t]{0.47\linewidth}
\centering
\includegraphics[width=4.1cm]{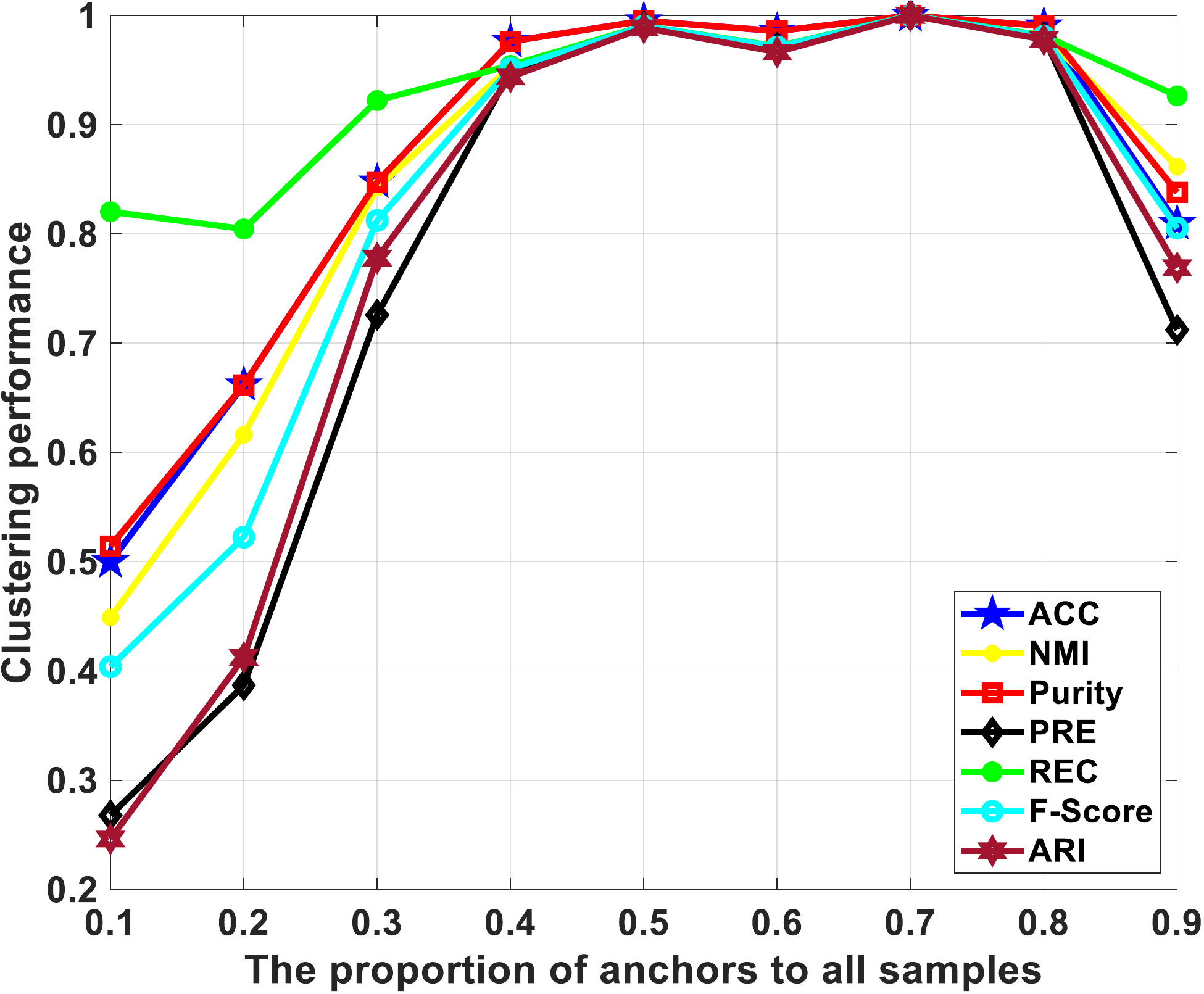}
\end{minipage}
}
\centering
\subfigure[Handwritten4]{
\begin{minipage}[t]{0.47\linewidth}
\centering
\includegraphics[width=4.1cm]{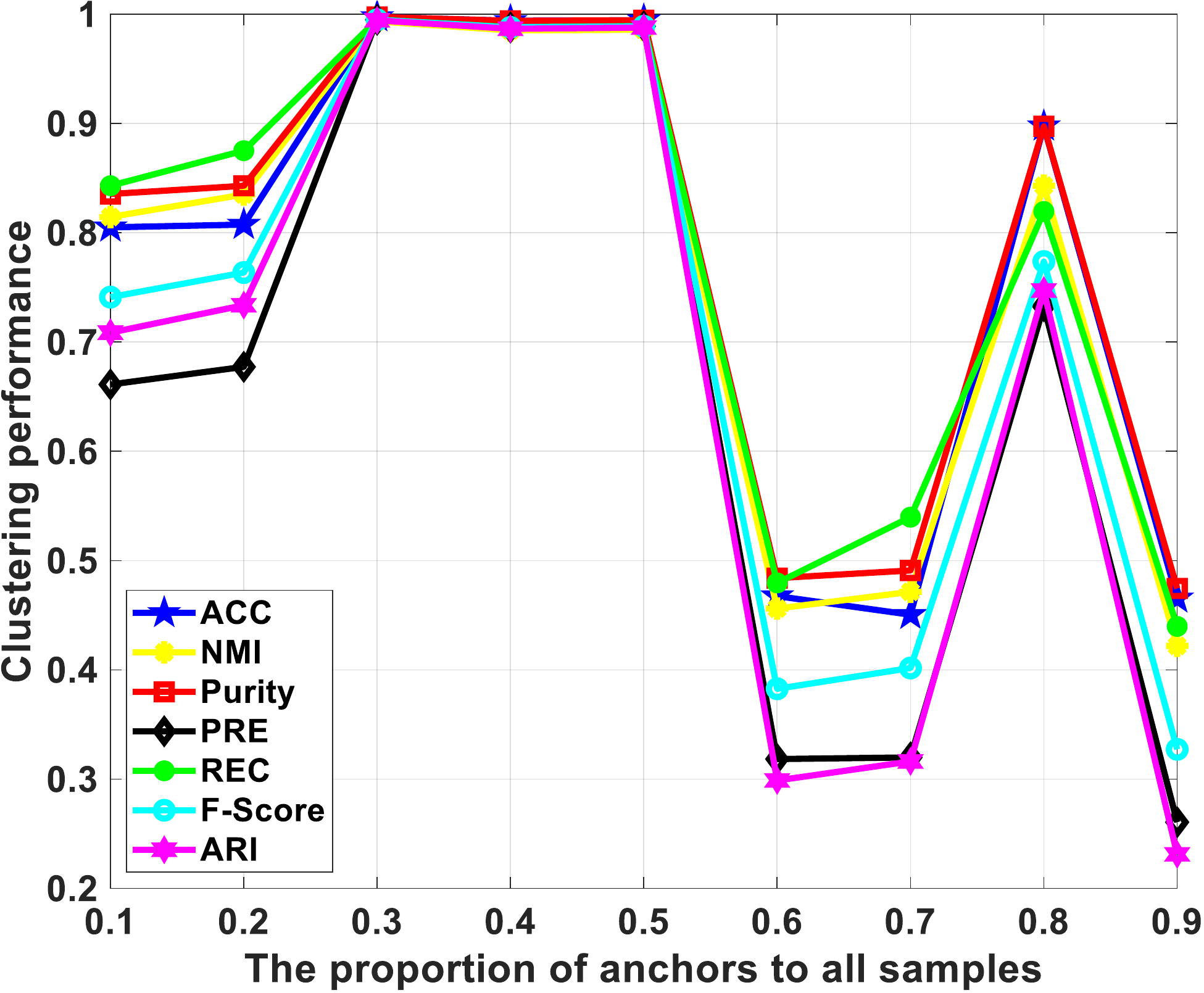}
\end{minipage}
}
\centering
\caption{The performances of our method with varying the number of anchor points on MSRC-v5 and Handwritten4 datasets.}
\label{a-value}
\vspace{-2mm}
\end{figure}

\begin{figure}[!t]
\begin{center}
\includegraphics[width=1.0\linewidth]{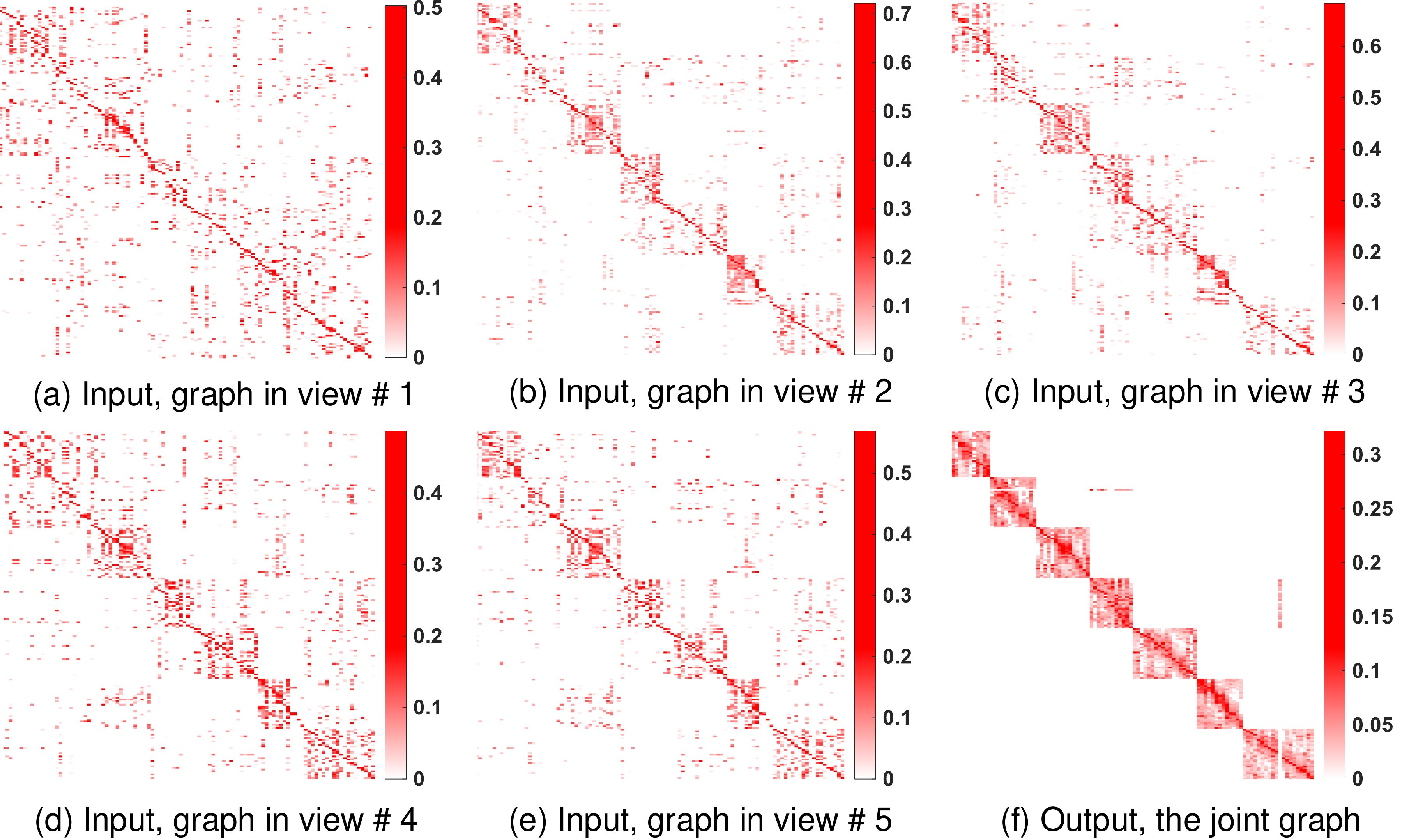}
\end{center}
   \caption{The graphs visualizations on MSRC-v5 dataset.}
\label{visual}
\vspace{-4mm}
\end{figure}
\subsection{Further Evaluation}
\textbf{Effect of parameter $p$:}
Taking MSRC-v5 and Handwritten4 datasets as examples, we analyze the impact of $p$ for clustering. Specifically, we change $p$ from $0.1$ to $1.0$ with the interval of $0.1$, then we report the ACC, NMI and Purity, as shown in Fig.~\ref{p-value}. One can observe that the results under different $p$ are distinguishing mostly, and when $p=0.9$ and $p=0.6$, we obtain the best clustering results on MSRC-v5 and Handwritten4 dataset, respectively. This demonstrates that $p$ has a significant influence on the clustering results. This is probably because that $p$ exploits the significant difference between singular values. Another reason may be that tensor Schatten $p$-norm makes the rank of the learned view-consensus graph approximate the target rank well.

\textbf{Effect of the number of anchors:}
We empirically analyze the effect of the number of anchors for clustering on MSRC-v5 and Handwritten4 datasets. To this end, we turn the proportion that anchors take in the entire data points from $0.1$ to $0.9$ with the interval of $0.1$, then we show seven metrics (ACC, NMI, Purity, PRE, REC, F-score and ARI) in Fig.~\ref{a-value}. It is clearly observed that our method has a large fluctuation with varying the number of anchors. When the proportion is set to $0.7$, our method obtains the best performance on MSRC-v5 dataset, and when the proportion is set to $0.3$, our method obtains the best performance on Handwritten4 dataset. Moreover, we find that the metrics curves \wrt anchors proportion are not monotonously increasing. This indicates that it is not necessary to use numerous anchors for clustering. Therefore, we set the anchors proportion to all data points as $0.5$ uniformly in the experiments.

\textbf{Graph Visualization:}
We present the input graphs and the learned view-consensus graph of our method on MSRC-v5 dataset in Fig.~\ref{visual}, where (a) - (e) are input graphs corresponding to five views, (f) is the view-consensus graph. It can be seen that the connected components in the input graphs of all five views are not clear. By employing our proposed method, we can observe that the learned view-consensus graph has exact $7$-connected components. It indicates that our method well characterizes the cluster structure. The above experimental results once again demonstrate that our proposed tensor Schatten $p$-norm minimization helps to ensure the rank of the learned view-consensus graph more similar to the target rank.

\section{Conclusion}
In this paper, we propose an effective and efficient graph learning for multi-view clustering. Our method learns the graph by minimizing our proposed tensor Schatten $p$-norm, which well characterizes the spatial structure and the complementary information embedded in different views. Different from most existing approaches, our method is time-economical due to the computation of the $n\times m$ ($m\ll n$) graph rather than $n\times n$, where $n$ and $m$ are the number of data points and anchors, respectively. By employing the a connectivity constraint, our method can directly obtain $K$-connected components. Extensive experiments on real-world datasets indicate that the efficiency of our method outperforms the state-of-the-art competitors.
\vspace{-2mm}
\section*{Acknowledgments}
The authors would like to thank the anonymous reviewers and AE for their constructive comments and suggestions.
\ifCLASSOPTIONcompsoc
\else
\fi
\ifCLASSOPTIONcaptionsoff
  \newpage
\fi
\vspace{-2mm}
\bibliographystyle{IEEEtran}
\bibliography{egbib}

\begin{thebibliography}{10}
\providecommand{\url}[1]{#1}
\csname url@samestyle\endcsname
\providecommand{\newblock}{\relax}
\providecommand{\bibinfo}[2]{#2}
\providecommand{\BIBentrySTDinterwordspacing}{\spaceskip=0pt\relax}
\providecommand{\BIBentryALTinterwordstretchfactor}{4}
\providecommand{\BIBentryALTinterwordspacing}{\spaceskip=\fontdimen2\font plus
\BIBentryALTinterwordstretchfactor\fontdimen3\font minus
  \fontdimen4\font\relax}
\providecommand{\BIBforeignlanguage}[2]{{%
\expandafter\ifx\csname l@#1\endcsname\relax
\typeout{** WARNING: IEEEtran.bst: No hyphenation pattern has been}%
\typeout{** loaded for the language `#1'. Using the pattern for}%
\typeout{** the default language instead.}%
\else
\language=\csname l@#1\endcsname
\fi
#2}}
\providecommand{\BIBdecl}{\relax}
\BIBdecl

\bibitem{Our}
Q.~Gao, W.~Xia, Z.~Wan, D.~Xie, and P.~Zhang, ``Tensor-svd based graph learning
  for multi-view subspace clustering,'' in \emph{AAAI}, 2020, pp. 3930--3937.

\bibitem{XXGHXG}
D.~Xie, X.~Zhang, Q.~Gao, J.~Han, S.~Xiao, and X.~Gao, ``Multiview clustering
  by joint latent representation and similarity learning,'' \emph{IEEE TC},
  vol.~50, no.~11, pp. 4848--4854, 2020.

\bibitem{ZhangFHCXTX20}
C.~Zhang, H.~Fu, Q.~Hu, X.~Cao, Y.~Xie, D.~Tao, and D.~Xu, ``Generalized latent
  multi-view subspace clustering,'' \emph{IEEE TPAMI}, vol.~42, no.~1, pp.
  86--99, 2020.

\bibitem{Jie2020}
J.~Wen, Y.~Xu, and H.~Liu, ``Incomplete multiview spectral clustering with
  adaptive graph learning,'' \emph{IEEE TC}, vol.~50, no.~4, pp. 1418--1429,
  2020.

\bibitem{NieLL16}
F.~Nie, J.~Li, and X.~Li, ``Parameter-free auto-weighted multiple graph
  learning: {A} framework for multiview clustering and semi-supervised
  classification,'' in \emph{IJCAI}, 2016, pp. 1881--1887.

\bibitem{ZhanZGW18}
K.~Zhan, C.~Zhang, J.~Guan, and J.~Wang, ``Graph learning for multiview
  clustering,'' \emph{IEEE TC}, vol.~48, no.~10, pp. 2887--2895, 2018.

\bibitem{HuNWL20}
Z.~Hu, F.~Nie, R.~Wang, and X.~Li, ``Multi-view spectral clustering via
  integrating nonnegative embedding and spectral embedding,'' \emph{Inf.
  Fusion}, vol.~55, pp. 251--259, 2020.

\bibitem{KumarD11}
A.~Kumar and H.~D. III, ``A co-training approach for multi-view spectral
  clustering,'' in \emph{ICML}, 2011, pp. 393--400.

\bibitem{KumarRD11r}
A.~Kumar, P.~Rai, and H.~D. III, ``Co-regularized multi-view spectral
  clustering,'' in \emph{NeurIPS}, 2011, pp. 1413--1421.

\bibitem{LiNHH15}
Y.~Li, F.~Nie, H.~Huang, and J.~Huang, ``Large-scale multi-view spectral
  clustering via bipartite graph,'' in \emph{AAAI}, 2015, pp. 2750--2756.

\bibitem{NieCLL18}
F.~Nie, G.~Cai, J.~Li, and X.~Li, ``Auto-weighted multi-view learning for image
  clustering and semi-supervised classification,'' \emph{IEEE TIP}, vol.~27,
  no.~3, pp. 1501--1511, 2018.

\bibitem{SFMC}
X.~Li, H.~Zhang, R.~Wang, and F.~Nie, ``Multi-view clustering: A scalable and
  parameter-free bipartite graph fusion method,'' \emph{IEEE TPAMI}, doi:
  {10.1109/TPAMI.2020.3011148}, 2020.

\bibitem{WuLZ19}
J.~Wu, Z.~Lin, and H.~Zha, ``Essential tensor learning for multi-view spectral
  clustering,'' \emph{IEEE TIP}, vol.~28, no.~12, pp. 5910--5922, 2019.

\bibitem{XuHNL17}
J.~Xu, J.~Han, F.~Nie, and X.~Li, ``Re-weighted discriminatively embedded
  k-means for multi-view clustering,'' \emph{IEEE TIP}, vol.~26, no.~6, pp.
  3016--3027, 2017.

\bibitem{LiuHC10}
W.~Liu, J.~He, and S.~Chang, ``Large graph construction for scalable
  semi-supervised learning,'' in \emph{ICML}, 2010, pp. 679--686.

\bibitem{NieHD12}
F.~Nie, H.~Huang, and C.~H.~Q. Ding, ``Low-rank matrix recovery via efficient
  schatten p-norm minimization,'' in \emph{AAAI}, 2012, pp. 665--661.

\bibitem{LL1}
F.~R. Chung and F.~C. Graham, \emph{Spectral graph theory}.\hskip 1em plus
  0.5em minus 0.4em\relax American Mathematical Soc., 1997, no.~92.

\bibitem{KFF}
K.~Fan, ``On a theorem of weyl concerning eigenvalues of linear transformations
  i,'' \emph{Proc. Natl. Acad. Sci. {USA}}, vol.~35, no.~11, pp. 652--655,
  1949.

\bibitem{LinLS11}
Z.~Lin, R.~Liu, and Z.~Su, ``Linearized alternating direction method with
  adaptive penalty for low-rank representation,'' in \emph{NeurIPS}, 2011, pp.
  612--620.

\bibitem{NieWDH17}
F.~Nie, X.~Wang, C.~Deng, and H.~Huang, ``Learning {A} structured optimal
  bipartite graph for co-clustering,'' in \emph{NeurIPS}, 2017, pp. 4129--4138.

\bibitem{NieWJH16}
F.~Nie, X.~Wang, M.~I. Jordan, and H.~Huang, ``The constrained laplacian rank
  algorithm for graph-based clustering,'' in \emph{AAAI}, 2016, pp. 1969--1976.

\bibitem{HaleYZ08}
E.~T. Hale, W.~Yin, and Y.~Zhang, ``Fixed-point continuation for
  l\({}_{\mbox{1}}\)-minimization: Methodology and convergence,'' \emph{{SIAM}
  J. Optim.}, vol.~19, no.~3, pp. 1107--1130, 2008.

\bibitem{TPAMI}
Q.~Gao, P.~Zhang, W.~Xia, X.~Deyan, X.~Gao, and D.~Tao, ``Enhanced tensor rpca
  and its application,'' \emph{IEEE TPAMI}, doi:{10.1109/TPAMI.2020.3017672},
  2020.

\bibitem{WinnJ05}
J.~M. Winn and N.~Jojic, ``{LOCUS:} learning object classes with unsupervised
  segmentation,'' in \emph{ICCV}, 2005, pp. 756--763.

\bibitem{Dua-2019}
\BIBentryALTinterwordspacing
D.~Dua and C.~Graff, ``{UCI} machine learning repository,'' 2017. [Online].
  Available: \url{http://archive.ics.uci.edu/ml}
\BIBentrySTDinterwordspacing

\bibitem{Deng12}
L.~Deng, ``The {MNIST} database of handwritten digit images for machine
  learning research [best of the web],'' \emph{{IEEE} Signal Process. Mag.},
  vol.~29, no.~6, pp. 141--142, 2012.

\bibitem{Fei-FeiFP07}
F.~Li, R.~Fergus, and P.~Perona, ``Learning generative visual models from few
  training examples: An incremental bayesian approach tested on 101 object
  categories,'' \emph{Comput. Vis. Image Underst.}, vol. 106, no.~1, pp.
  59--70, 2007.

\bibitem{ChuaTHLLZ09}
T.~Chua, J.~Tang, R.~Hong, H.~Li, Z.~Luo, and Y.~Zheng, ``{NUS-WIDE:} a
  real-world web image database from national university of singapore,'' in
  \emph{{ACM} {CIVR}}, 2009.

\bibitem{ApteDW94}
C.~Apt{\'{e}}, F.~Damerau, and S.~M. Weiss, ``Automated learning of decision
  rules for text categorization,'' \emph{{ACM} Trans. Inf. Syst.}, vol.~12,
  no.~3, pp. 233--251, 1994.

\bibitem{NieLL17}
F.~Nie, J.~Li, and X.~Li, ``Self-weighted multiview clustering with multiple
  graphs,'' in \emph{IJCAI}, 2017, pp. 2564--2570.

\bibitem{KangZZSHX20}
Z.~Kang, W.~Zhou, Z.~Zhao, J.~Shao, M.~Han, and Z.~Xu, ``Large-scale multi-view
  subspace clustering in linear time,'' in \emph{AAAI}, 2020, pp. 4412--4419.

\bibitem{xie2018hyper}
Y.~Xie, W.~Zhang, Y.~Qu, L.~Dai, and D.~Tao, ``Hyper-laplacian regularized
  multilinear multiview self-representations for clustering and semisupervised
  learning,'' \emph{IEEE TC}, vol.~50, no.~2, pp. 572--586, 2020.

\end{thebibliography}

\end{document}